# The Multiplex Classification Framework: optimizing multi-label classifiers through problem transformation, ontology engineering, and model ensembling

Mauro Nievas Offidani[a,*], Facundo Roffet[a,b], Claudio Augusto Delrieux[a,b], María Carolina González Galtier[c], and Marcos Zarate[d]

[a]*Department of Electric and Computer Engineering, Universidad Nacional del Sur, Bahía Blanca, Argentina*

[b]*Institute of Computer Science and Engineering, National Scientific and Technological Research Council of Argentina (CONICET)*

[c]*Freelance Healthcare Data Analyst*

[d]*Centre for the Study of Marine Systems, Centro Nacional Patagónico (CENPAT-CONICET), Puerto Madryn, Argentina*

**Abstract**. Classification is a fundamental task in machine learning. While conventional methods—such as binary, multiclass, and multi-label classification—are effective for simpler problems, they may not adequately address the complexities of some real-world scenarios. This paper introduces the Multiplex Classification Framework, a novel approach developed to tackle these and similar challenges through the integration of problem transformation, ontology engineering, and model ensembling. The framework offers several advantages, including adaptability to any number of classes and logical constraints, an innovative method for managing class imbalance, the elimination of confidence threshold selection, and a modular structure. Two experiments were conducted to compare the performance of conventional classification models with the Multiplex approach. Our results demonstrate that the Multiplex approach can improve classification performance significantly (up to 10% gain in overall F1 score), particularly in classification problems with a large number of classes and pronounced class imbalances. However, it also has limitations, as it requires a thorough understanding of the problem domain and some experience with ontology engineering, and it involves training multiple models, which can make the whole process more intricate. Overall, this methodology provides a valuable tool for researchers and practitioners dealing with complex classification problems in machine learning.

* Corresponding author. E-mail: mauro.offidani@uns.edu.ar.

## 1. Introduction

### *1.1 Purpose*

Classification problems, one of the main tasks in machine learning (ML), are a type of supervised learning task in which the goal is to assign classes to new objects or instances given a set of possible classes and considering past observations (training data). Classic examples of classification problems include detecting spam emails, identifying handwritten characters, and labeling movies according to their content by using binary, multiclass, and multi-label classifiers, respectively. However, our experience with medical image classification has shown us that real-world scenarios can sometimes involve much greater complexity, and therefore may not be adequately addressed by conventional classification methodologies. A classification problem is considered to be complex when it involves a large number of potential classes and intricate logical relationships between them, such as hierarchical dependencies, overlapping categories, and mutual exclusivity constraints. This leads us to our primary research question: what is the most effective way to approach complex classification problems?

The aim of this article is to present a classification approach capable of handling problems with any number of classes and logical constraints between them. In addition, the contributions of this paper include: a) a methodology to adapt any classification problem by applying problem transformation, ontology engineering, and model ensembling; b) the corresponding code and practical examples; c) evidence of the framework's impact through empirical results; d) a method to improve data quality by utilizing logical constraints among classes; and e) a novel approach to addressing class imbalance problems.

Over the course of this paper, the main types of classification problems in machine learning will be defined (Section 1.2) and related work will be reviewed, focusing on existing approaches to handling complex classification problems and ontology-based solutions (Section 2). Then, the fundamental concepts of the Multiplex Classification Framework will be introduced, including compound classes, basic classification tasks, decision rainforests, and divergent cascading ensembles (Section 3). Following this, the main steps of the framework will be explained (Section 4). Two experiments that were performed to validate this framework will then be described (Section 5), and their results will be analyzed (Section 6). Finally, the implications of the findings will be discussed (Section 7), and the key contributions will be summarized in the conclusion (Section 8).

### *1.2 Main classification problems in Machine Learning*

Classification problems can be categorized based on the number of possible classes, the amount of possible outcomes per instance, and the presence of multiple parallel or sequential steps (Fig. 1). For the purposes of this paper, each individual step in a classification problem will be referred to as a 'classification task.' It is important to note that this article will not address the ML algorithms used to solve classification problems, such as support vector machines, k-nearest neighbors, or neural networks.

The main types of classification problems in ML are:

- **Binary Classification:** Instances are assigned to one and only one class from a set of two possible classes. An example of this is a spam classifier, which is used to categorize emails as either *spam* or *nonspam* (Drucker et al., 1999).
- **Multiclass Classification:** Instances are assigned to one and only one class from a set of more than two possible classes. One of the best-known datasets for this type of task is the MNIST dataset, used to train models that recognize handwritten digits by classifying images into one of ten possible digits (LeCun et al., 1998).

- **Multi-Label Classification:** Instances can be assigned to any number of classes (including none) from a set of two or more classes. This type of model is used, for example, by streaming platforms to assign multiple tags to their content (Akbar et al., 2022).
- **Multitask Classification:** This approach involves training a single model to perform multiple classification tasks simultaneously, each with its own set of possible classes. An example of this is a facial expression analysis model that simultaneously detects smiles, recognizes emotions, and classifies gender (Sang et al., 2017).
- **Multilevel or Hierarchical Classification:** This type of model performs multiple classification tasks sequentially, with the set of possible classes at a given step depending on the class assigned in the previous step. An example of this is the classification of living organisms into different levels of the taxonomic hierarchy (depending on the Kingdom, Phylum, Class, Order, Family, Genus, and Species) (Gomez-Donoso et al., 2021).
- **Multiplex Classification:** This approach, introduced in this paper, is designed to address classification problems that involve both sequential and simultaneous tasks. Figure 2 shows a simplified classification problem from the medical imaging domain: initially, all images are classified by their type, and then, images classified as *ultrasound* are further categorized based on two independent criteria: the site of the image and whether Doppler is used.

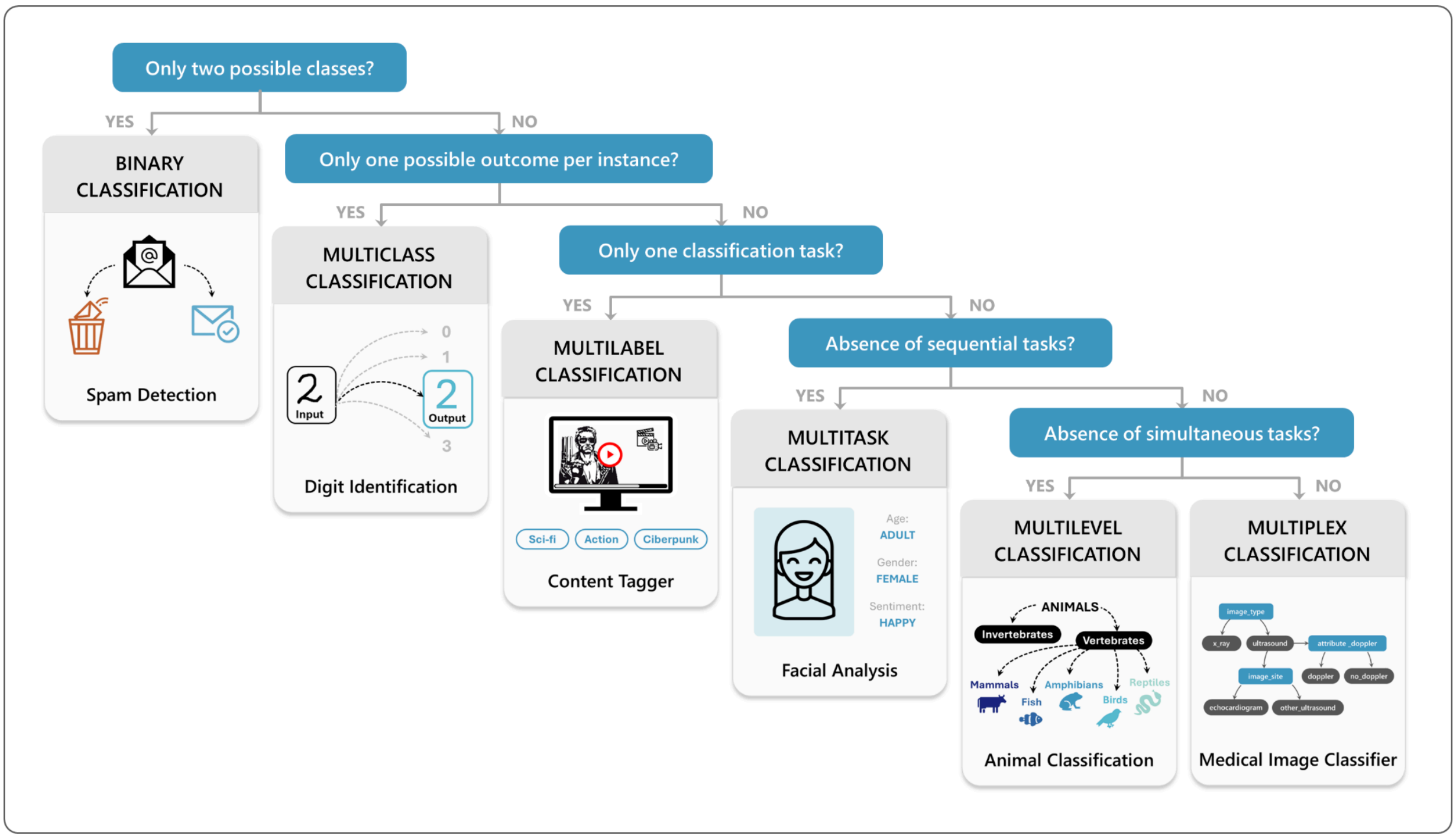

Fig. 1. Main types of classification problems and examples.

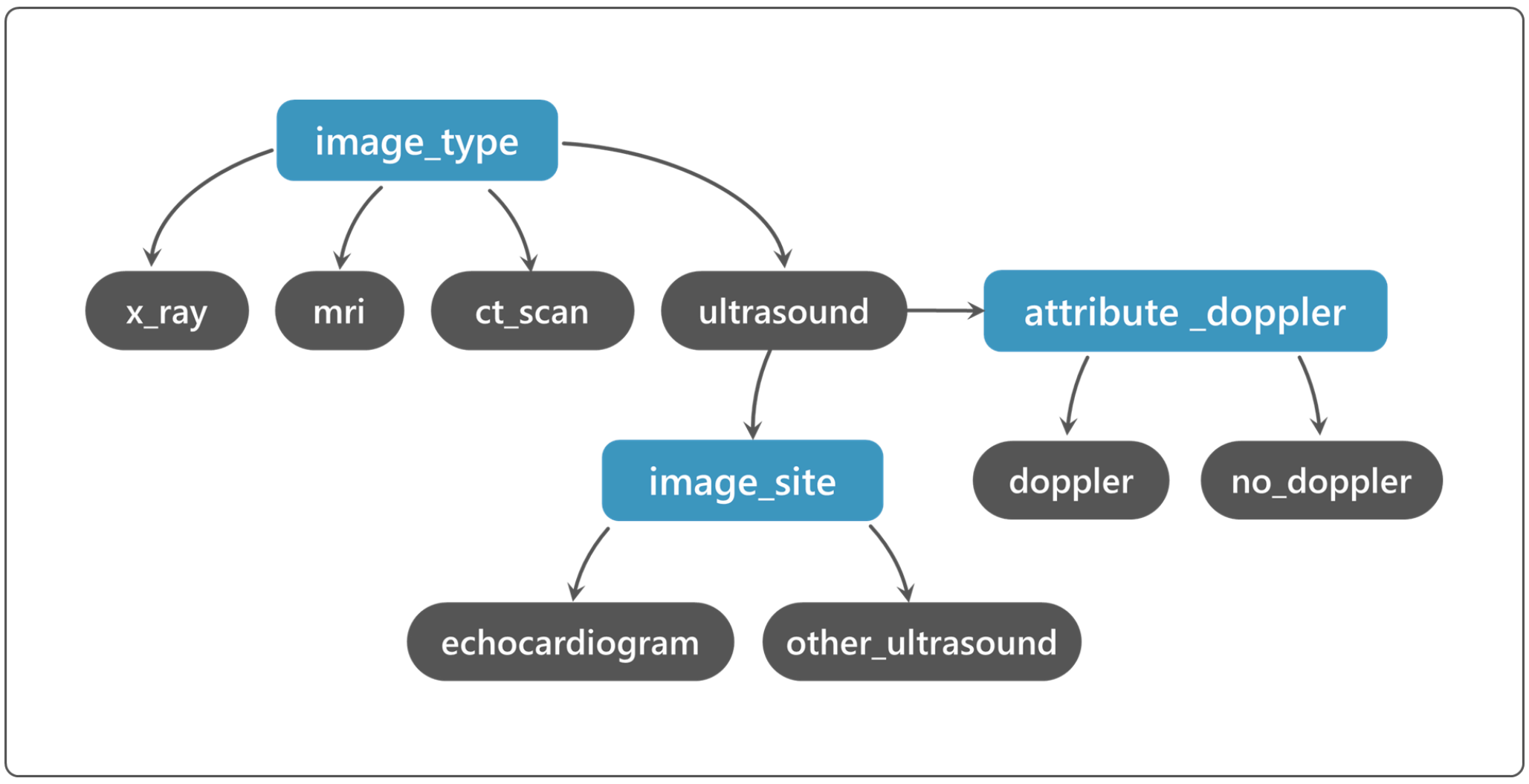

Fig. 2. Example of a Multiplex Classification problem.

## 2. Related work

### *2.1 Approaches to complex classification problems*

Besides the Multiplex Classification Framework, other methodologies have been developed to deal with complex classification problems present in real-world scenarios, namely Extreme Multi-Label Classification, Hierarchical Multi-Label Classification, and Directed Acyclic Graphs. Each of these methods has its strengths and limitations and, as will be explained in Section 3.2, they can all be transformed into a Multiplex Classification problem.

#### 2.1.1 Extreme Multi-Label Classification

The Extreme Multi-Label Classification (XMLC) is an approach that focuses on multi-label classification problems with an extremely large number of labels (Liu et al., 2022). A typical example of an XMLC application is a facial recognition system that identifies individuals in a given image from a set of millions of possible people. Useful resources on this approach can be found in the Extreme Classification Repository, which includes datasets with up to nearly 3 million labels (Bhatia et al., 2016).

Although XMLC is a useful method for addressing classification tasks with thousands or millions of classes, it does not fundamentally differ from multi-label classification in terms of how it organizes the set of possible classes, as there are no hierarchical relations or logical constraints among them. Tree-based methods, a key XMLC technique, partition the label set hierarchically, dividing the classification problem into a sequence of tasks. However, these divisions are not based on logical relationships between classes but on the feature vectors of the input data and the distribution of labels in the training instances (Khandagale et al., 2020; Liu et al., 2022; Prabhu & Varma, 2014).

### 2.1.2 Hierarchical Multi-Label Classification

In Hierarchical Multi-Label Classification (HMC), classes are hierarchically structured and instances may belong to multiple classes at the same hierarchical level (Cerri et al., 2011; Vens et al., 2008). This approach admits only one parent class per class. An example of HMC is the use of ICD codes to classify Electronic Health Records (EHRs). ICD codes are structured hierarchically, with broader categories (e.g., *Diseases of the respiratory system*) and more specific subcategories (e.g., *Asthma*), and multiple codes can be assigned to the same patient (World Health Organization, 2004). A key limitation of HMC is that it does not contemplate the possibility of mutual exclusivity between classes that share the same parent class.

### 2.1.3 Directed Acyclic Graphs

In Directed Acyclic Graphs (DAGs) there are hierarchical relations between classes, and instances can belong to different classes at the same hierarchical level just like in HMC (Ramírez-Corona et al., 2014). The main difference with HMC is that classes can have multiple parents as long as no cycle is formed (that means, for example, that a child class cannot be an ancestor of its own parent class) (Fig. 3). This kind of approach is used, for example, to make predictions based on DAG-structured ontologies, such as the Gene Ontology or the Human Phenotype Ontology (Notaro et al., 2017, 2021; Valentini, 2014). This approach allows more complex relations between classes, but it still has some limitations: mutual exclusivity between sibling classes is not contemplated, and the resulting structure may be too intricate and confusing when there is a relatively large number of classes (Fig. 4).

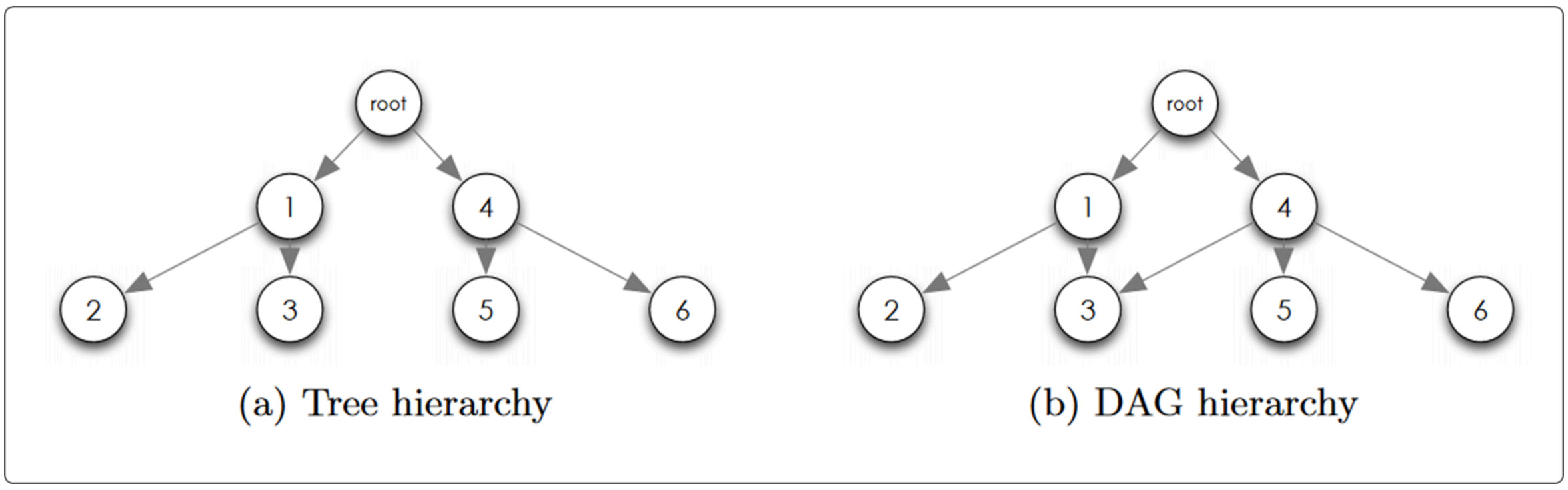

Fig. 3. Difference between tree and DAG structures, taken from Ramírez-Corona et al. (2014). Note that classes in DAGs can have more than one parent class.

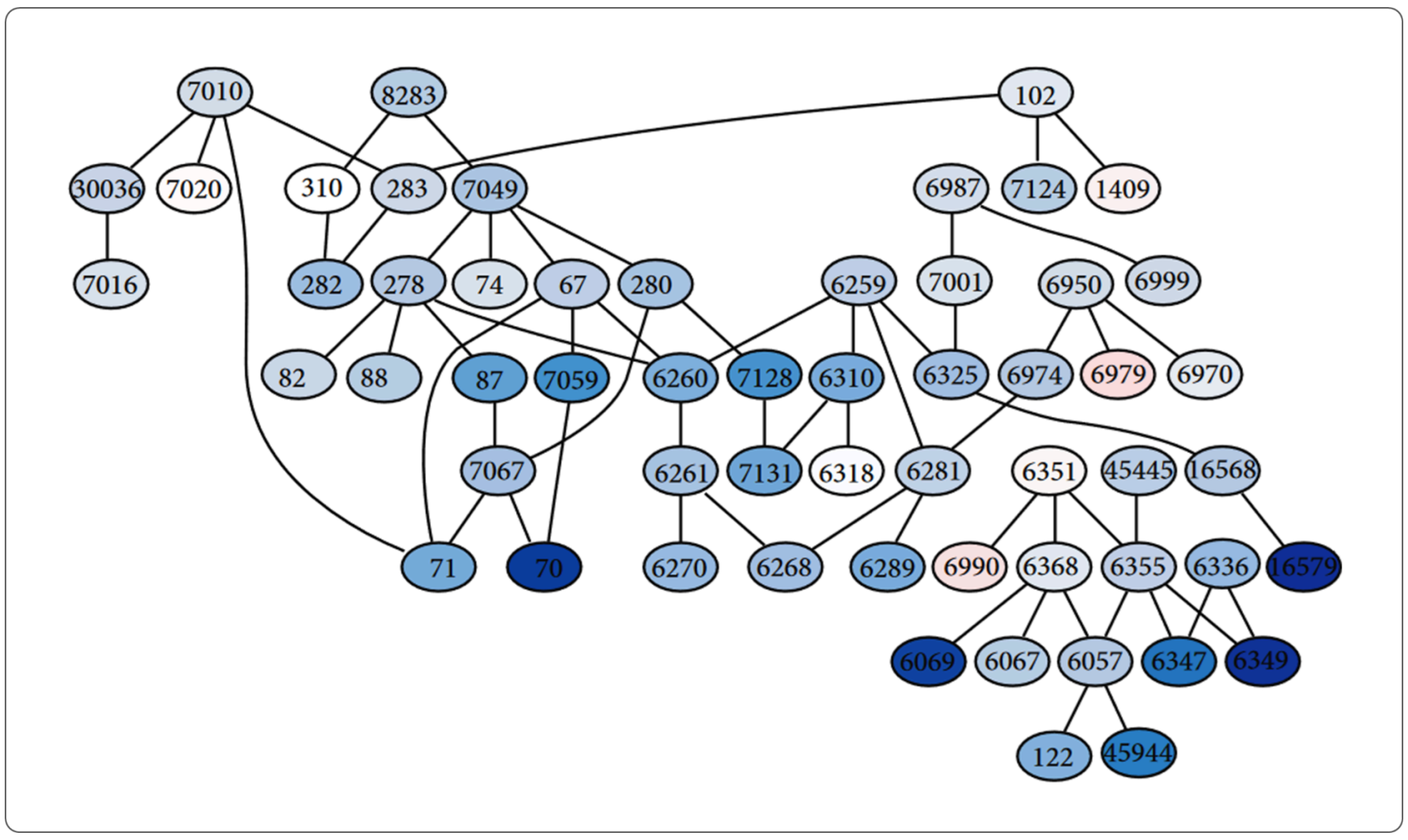

Fig. 4. A DAG based on part of the Gene Ontology (adapted from Valentini, 2014).

### *2.2 Ontology-based approaches to multi-label classification*

Many studies have leveraged domain ontologies, such as the Gene Ontology, to enhance the performance of multi-label classification models (Jin et al., 2008; Mazo et al., 2020; Vogrincic & Bosnic, 2011; Zhou & El-Gohary, 2015). One advantage of these ontology-based approaches, also evident in the Multiplex Classification Framework, is their ability to apply logical constraints to predicted classes, enabling the removal of incompatible labels and similar actions to improve model performance. Additional documented benefits of these approaches, compared to traditional multi-label classification, include addressing data imbalance by performing one-vs.-rest classification locally rather than globally (Jin et al., 2008), and enabling the use of unsupervised learning techniques (which do not require annotated data), such as calculating the similarity between ontology terms and the contents of the document for text classification (Zhou & El-Gohary, 2015). However, these approaches typically involve multi-label classifiers, multi-level classifiers, hierarchical multi-label classifiers, or DAGs, and therefore face the limitations mentioned in Section 2.1. It is also important to note that, in many studies, classes and relations in domain ontologies are used as-is, without adaptation, based on the assumption that human-constructed ontologies are inherently suitable for ML tasks (something that may not always be the case, as will be demonstrated in Section 4.1.5).

## 3. Fundamental concepts

In this section, four concepts that are essential to the Multiplex Classification Framework will be defined and characterized: *compound classes* (classes that combine multiple concepts), *basic classification tasks* (the fundamental elements that make up any classification problem), *decision rainforests* (the structure and organization of these elements within a classification problem), and *divergent cascading ensembles* (a methodology for assembling ML models that mimics the structure of decision rainforests). Section 3.5 includes a demonstration of how these concepts can be applied to different types of classification problems.

### *3.1 Compound classes*

The term *compound class* refers to individual classes that combine multiple concepts and therefore can be broken down into simpler classes. For example, the class *echocardiography* is a compound class that, in a given classification problem, could be split into the component classes *heart* and *ultrasound*. Compound classes are not included in the class set of Multiplex classification problems; instead, they are split into their component classes during data pre-processing and reintroduced during post-processing, as explained in Section 4.2. This approach is used to avoid graph structures (since component classes would have more than one parent class) and because compound classes do not involve any actual classification task: whenever an instance belongs to all component classes of a given compound class, it automatically belongs to that compound class.

### *3.2 Basic classification tasks*

Basic classification tasks (BCTs) are tasks in which the classes are mutually exclusive and collectively exhaustive. This means that any instance or object is always assigned one and only one class from a given set of classes. BCTs can be categorized based on the number of possible classes: binary (with only two classes) or multiclass (with more than two classes).

While some real-world classification problems may involve a single BCT, most involve a combination of them. BCTs can be classified as unconditional or conditional. Unconditional BCTs apply to all instances in a classification problem, whereas conditional BCTs are applied only if an instance belongs to a specific class from a prior BCT. For example, if a classification problem involves first classifying objects as *living_object* or *non_living_object*, and then further classifying *living_object* instances by species (e.g., *cat*, *dog*), the second task is considered a conditional BCT.

### *3.3 Decision rainforests*

Taxonomies belong to the ontology spectrum, which also includes artifacts such as formal upper-level ontologies and controlled vocabularies (Gruninger et al., 2008). They are primarily characterized by their hierarchical organization, and they employ a more restricted language than general ontologies, focusing mainly on specifying subclass relationships between classes. In this work, the word *taxonomy* refers to class sets in ML classification problems, whereas *flat taxonomy* is used for class sets that lack both hierarchical structure and logical constraints. The term *decision rainforest* is used to refer to tree-based taxonomy structures used for Multiplex classification problems. This approach enables the representation of any number of classes and class constraints without relying on complex structures such as graphs or advanced ontological concepts like properties or axioms. This is particularly useful given that domain experts involved in ontology development or adaptation may make errors or omissions when required to use formal languages or complex ontological concepts they are unfamiliar with (Westerinen & Tauber, 2017).

Decision rainforests are conformed by the classes from a given unconditional BCT and from any BCT that is conditioned by it (Fig. 5). The main tree contains the unconditional BCT and not more than one conditional BCT per class. If a given class has more than one conditional BCT, only one of them will belong to the main tree, and there will be a new tree for each of the rest of them. These new trees are called subsidiary trees and, just like the main tree, they can contain multiple BCTs and can also have their own subsidiary trees. The decision of which BCT belongs to the main tree and which ones form subsidiary trees is arbitrary (in other words, any BCT could be part of the main tree).

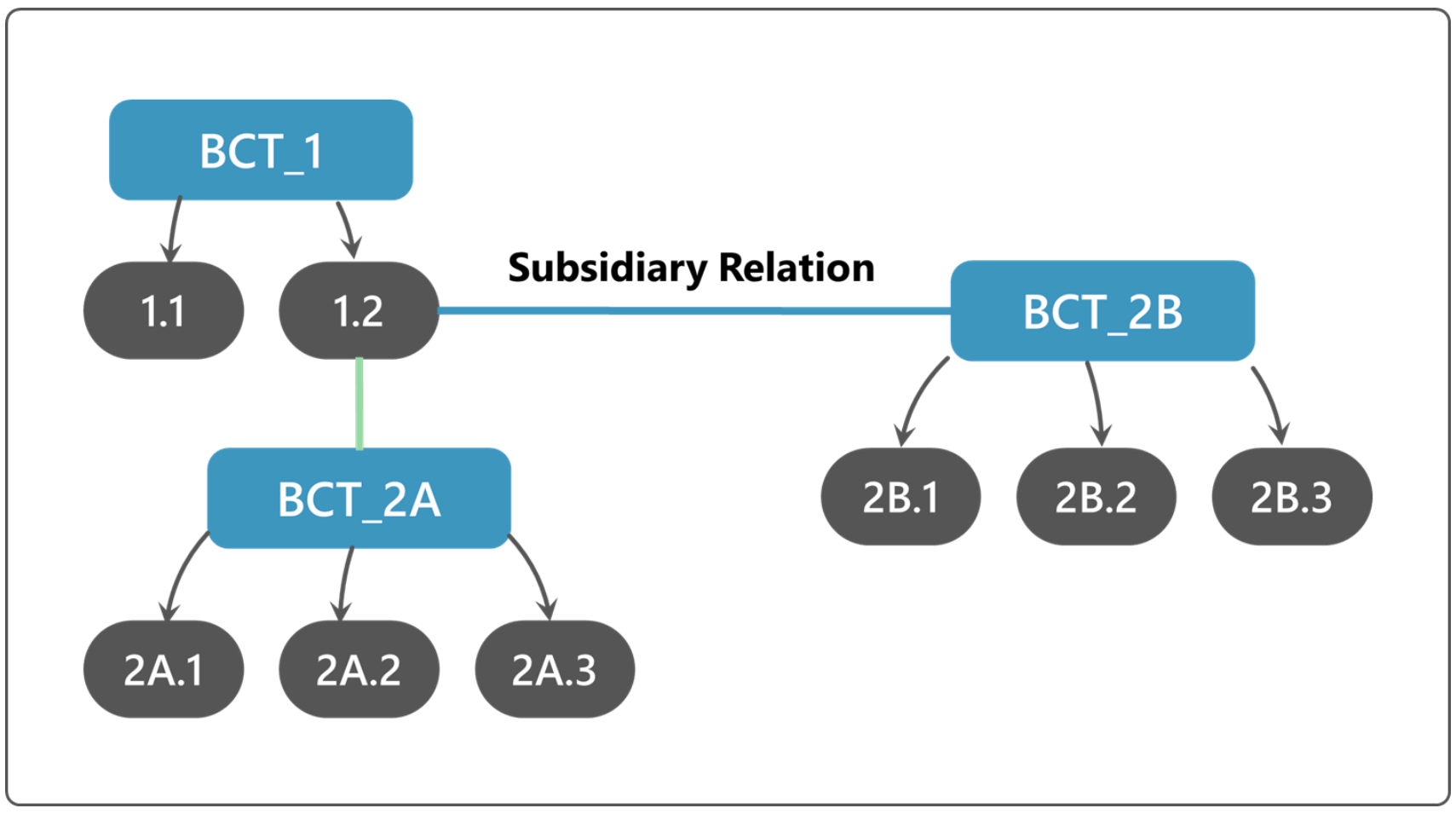

Fig. 5. Example of decision rainforest. Rectangles represent BCTs, and ovals represent classes. Each BCT is linked to its corresponding classes through arrows. Lines without arrowheads represent conditioning relations between a class and a BCT. The green line represents a hierarchical relation, and the blue line represents a subsidiary relation.

In the example from Fig. 5, BCT_1 is an unconditional BCT, and BCT_2A and BCT_2B are conditional to class *1.2*. All the classes from BCT_1 and BCT_2A belong to the same tree (main tree), and the classes from BCT_2B belong to a subsidiary tree. Conditioning relationships that belong to the same tree are called hierarchical relations, and the ones that link a class with a subsidiary tree are called subsidiary relations. This structure is called *decision rainforest* because it consists of decision trees that are interconnected through subsidiary relations that resemble lianas in a rainforest (in contrast to decision forests where the trees are independent).

### *3.4 Divergent cascading ensemble*

For Multiplex Classification, instead of using a single ML model, multiple models are assembled forming a hyper-architecture that mirrors the structure of the corresponding decision rainforest. This hyper-architecture is called divergent cascading ensemble because different models work in sequence (following hierarchical relations) and the process diverges whenever a predicted class has more than one conditional BCT. Multitask classifiers are used in the points of the process where it diverges, and binary or multiclass classifiers are used in the rest of the process. Multitask classifiers are employed due to their potential to enhance task-specific accuracy, compared to training the models separately (Caruana, 1997). In the example from Fig. 6, first a binary classifier is used for the unconditional BCT (BCT_1), and then a multitask classifier is used for the BCTs that are conditional to the class predicted in BCT_1 (BCT_2A and BCT_2B).

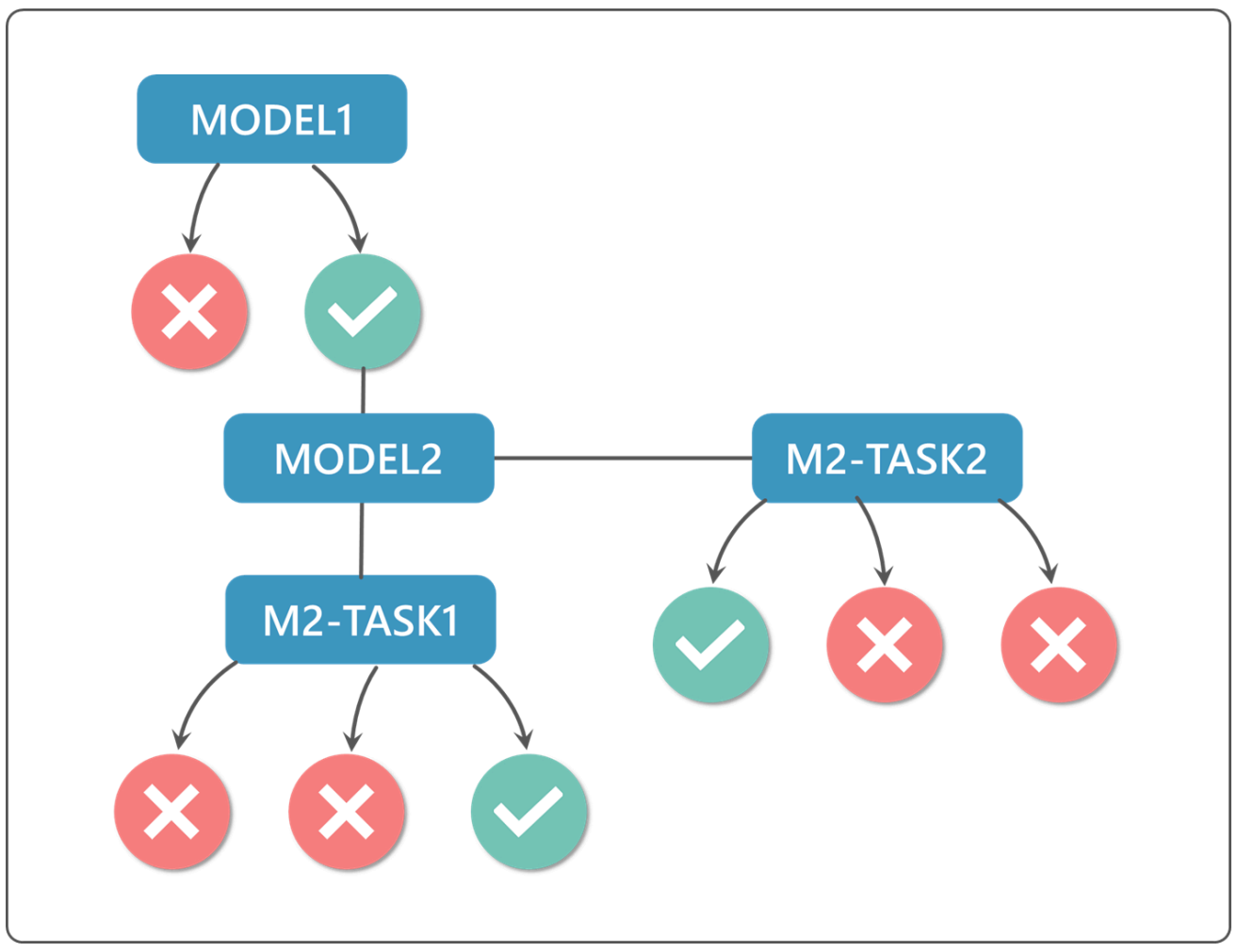

Fig. 6. Example of Divergent Cascading Ensemble. MODEL1 is a binary classifier, and MODEL2 is a multitask classifier involving two multiclass tasks. The ticks represent predicted classes, and the crosses represent discarded classes.

### *3.5 Implementation of fundamental concepts*

The aim of this section is to show that the core concepts of the Multiplex Classification Framework can be applied to any generic classification problem through a process known as problem transformation (the actual methodology will be detailed in the next section). In certain cases, this process involves the creation of auxiliary superclasses or exclusion classes, which are subsequently removed during postprocessing.

Problem transformation is very straightforward in some cases: **Binary Classification** is equivalent to a single binary BCT, **Multiclass Classification** is the same as a single multiclass BCT, and **Multi-Label Classification** can be turned into multiple simultaneous binary BCTs (one per label) following the Binary Relevance method (Cherman et al., 2011). For **Extreme Multi-Label Classification**, in addition to using the Binary Relevance method, it may be possible to transform the problem into a combination of sequential and simultaneous BCTs, each with a limited number of classes. Details on how to split a BCT can be found in Section 4.1.3.

When it comes to **Multitask Classification**, each task should be transformed independently: first, determine whether each task is binary, multiclass, or multi-label, and then transform them into BCTs according to the methods described previously. **Multilevel Classification** can be turned into a sequence of BCTs connected by hierarchical relations. In **Hierarchical Multi-Label Classification**, each hierarchy level should be turned into multiple simultaneous binary BCTs (one per each label in that level, just as in Multi-Label Classification), and a conditioning relation should be created between each binary BCT and the corresponding class from the previous level (only one of those relations is considered hierarchical, the rest of them are subsidiary relations).

Regarding **DAGs**, the logic is the same as in Hierarchical Multi-Label Classification except for classes with more than one parent class (something that is not possible in Multiplex Classification). In these cases, it is necessary to change the structure of relations between classes so that the class with multiple parents becomes part of a subsidiary tree from a class in an upper level of hierarchy. In the process, the corresponding auxiliary superclasses and negative classes are created (see example in Fig. 7). Although the resulting class structure may seem more complex than the input DAG, it replaces the use of graphs with simpler models and it provides the benefits of the Multiplex approach (see Section 7). It is important to note that this transformation process should only be performed once compound classes and wrong relationships have been ruled out, to avoid propagating errors from the input taxonomy. If compound classes are present, they should be split into their component classes (see Section 4.1.1), and any incorrect relationships between classes should be corrected or removed before proceeding with the transformation.

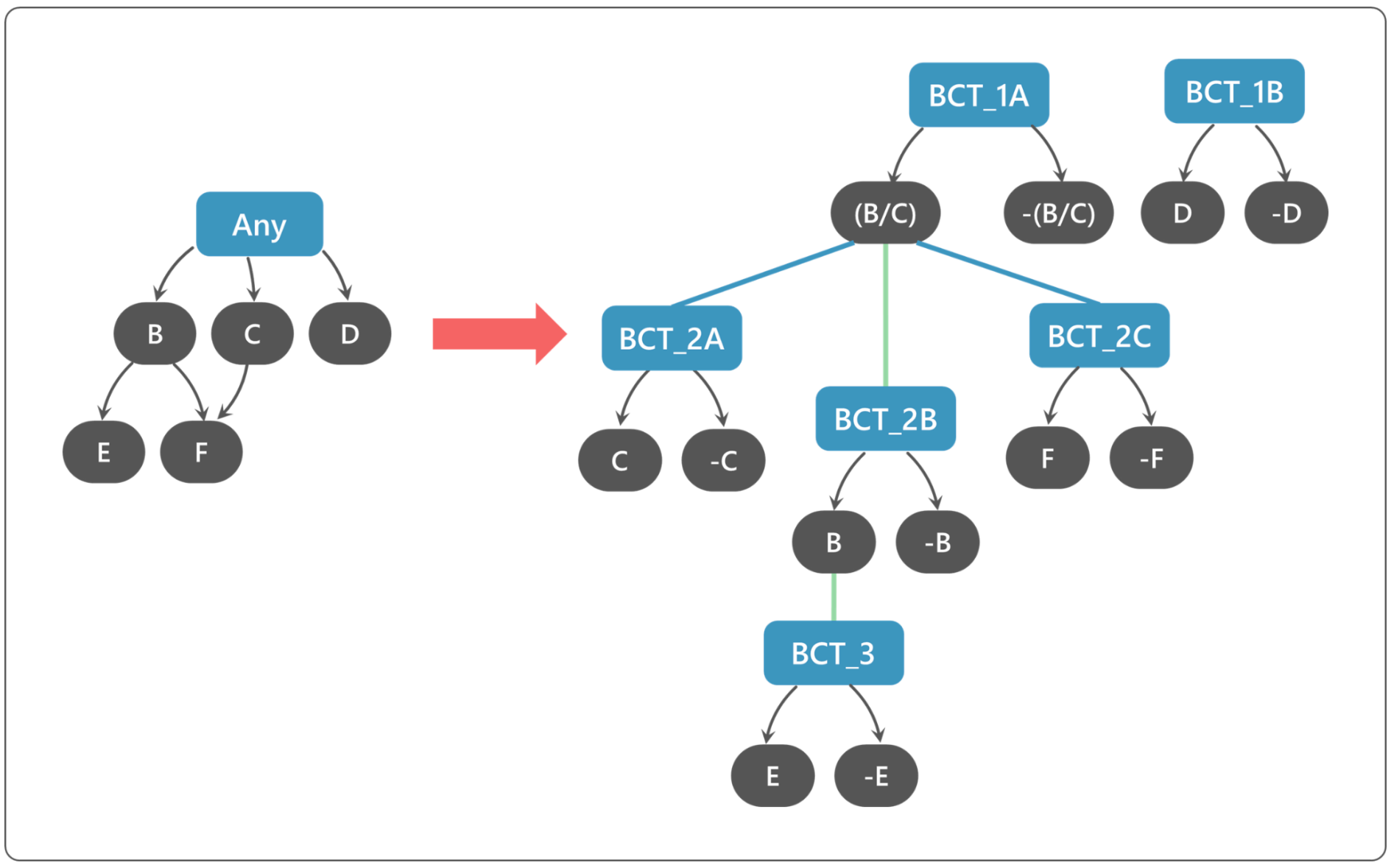

Fig. 7. Example of a generic DAG (left) transformed into a decision rainforest (right). *(B/C)* is an auxiliary superclass used for instances that belong to class *B* and/or to class *C*. The minus sign is used for negative classes (for instance, the class *-D* is used for any instance that does not belong to *D*). Each class is turned to binary BCTs because there are no constraints among classes from the same hierarchical level. Note that the class *F*, which has two parent classes in the DAG, is included in the decision rainforest in a subsidiary tree that is conditional to the class *(B/C)*.

## 4. The Multiplex Classification Framework

This section includes an explanation of how any specific classification problem can be transformed according to the Multiplex Classification Framework. The overall process consists of the following tasks:

1. Taxonomy Adaptation
    a. Class Decomposition
    b. Identification of Logical Constraints
    c. Taxonomy Structuring
    d. Creation of the OWL File

2. Data Preparation
3. Model Ensembling
    a. Model Training
    b. Model Inference

In this process, domain experts or ontologists handle taxonomy adaptation, while data scientists are responsible for data preparation and model ensembling. Code and examples can be found in the project's repository: https://github.com/mauro-nievoff/Multiplex_Classification.

### *4.1 Taxonomy adaptation*

The steps used to adapt any taxonomy incorporate elements from methods applied in the fields of Ontology Engineering and Ontology Quality Assurance (Amith et al., 2018; Noy & McGuinness, 2001). Taxonomy adaptation can be applied to both a class set from a given ML classification problem or to domain ontologies that were not originally intended for ML classification. An example of how to adapt a domain ontology is provided in Section 4.1.5.

#### 4.1.1 Class decomposition

First, all the classes from the initial taxonomy should be listed. If any compound class is present, then it should be broken down so that no concept appears in more than one class (e.g. if the initial taxonomy includes *ct_with_contrast* and *mri_with_contrast*, the resulting classes after splitting should be *ct*, *mri* and *contrast*). If the input domain ontology contains attributes associated with a given class, the values for those attributes should be treated as separate classes and included in the list (since, in ML, all problems involving categorical variables are addressed using classification models).

At this point, it’s important to take into account that sometimes class names can be misleading. For example, if an image is classified as both *tumor* and *fracture*, it might seem like an error because, in real life, something cannot be both a tumor and a fracture simultaneously. However, in this example, the class names *tumor* and *fracture* are actually short forms for *image_with_tumor* and *image_with_fracture*, respectively, and these classes are not mutually exclusive because an image can indeed contain both a tumor and a fracture at the same time. In contrast, if this were an object detection problem rather than an image classification problem, the classes *tumor* and *fracture* would be mutually exclusive, as the classification instances would be defined differently (with objects in the image being classified individually rather than classifying the entire image).

#### 4.1.2 Identification of logical constraints

After class decomposition, it is necessary to identify logical constraints between classes by analyzing all the possible pairs of classes (generically referred to as *Class_A* and *Class_B*). There are four possible scenarios (see Fig. 8):

- **Complete overlap:** if all instances from *Class_A* are instances from *Class_B* and vice versa, these classes should be merged into one to avoid redundancy (e.g. *x_ray* and *roentgenogram* should be both merged into *x_ray* because they are synonyms).
- **Class containment:** if *Class_A* contains *Class_B* (i.e. all instances from *Class_B* belong to *Class_A* but not all instances from *Class_A* belong to *Class_B*), then *Class_A* is either the direct parent of *Class_B* or any other of its ancestor classes. In other words, this means that *Class_A* conditions the BCT that includes *Class_B*. The classes *radiology* and *x_ray* illustrate this concept: all instances classified as *x_ray* belong to the class *radiology*, but not all the instances of *radiology* belong to the class *x_ray* (i.e., *radiology* contains *x_ray*).

- **Mutual exclusion:** if none of the instances from *Class_A* is an instance from *Class_B*, then it means that they belong to the same BCT (if they have the same parent class) or that *Class_A* or one of its ancestor classes belongs to the same BCT as *Class_B* or one of its ancestor classes. For example, consider the classes *ultrasound* and *x_ray*: they have the same parent class (*radiology*) and no instance belongs to both classes at the same time, which means that they belong to the same BCT.
- **Partial overlap:** if some (but not all) instances from *Class_A* are also instances from *Class_B* and vice versa, then it means that: a) they belong to different decision rainforests, or b) they belong to different trees from the same rainforest and they share an ancestor class. Consider the classes *ct* and *angiography* as an example: only some instances belong to both classes, while others belong only to either *ct* or *angiography*. This occurs because they share the same parent class (*radiology*) but belong to different BCTs and trees within the decision rainforest.

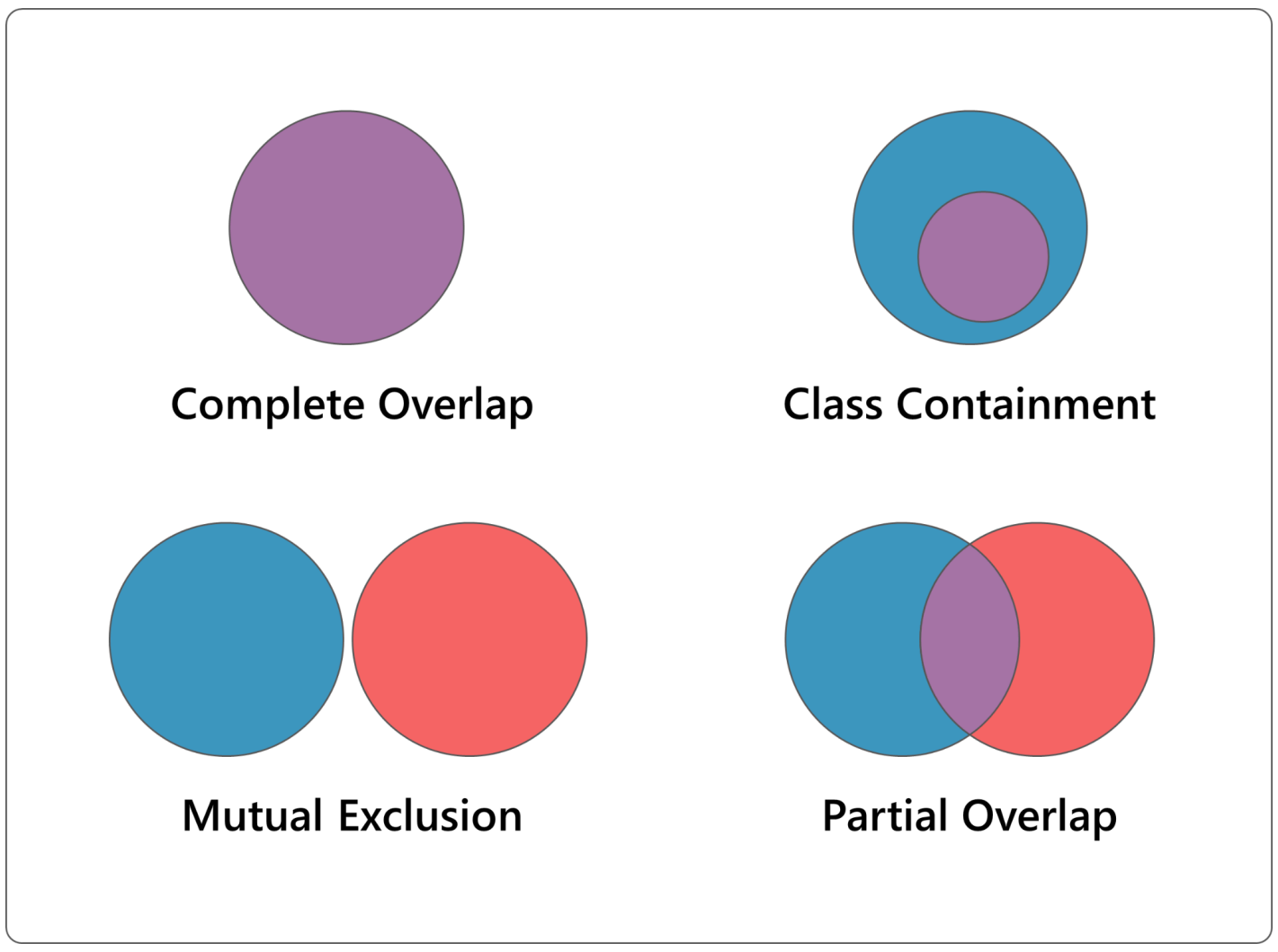

Fig. 8. Venn diagrams illustrating the possible logical relationships between two classes, represented by a red circle and a blue circle. The areas shaded in purple indicate the overlap between the classes.

4.1.3 Taxonomy structuring

Once all the logical constraints are identified, it is possible to start working on the structure of the taxonomy. First, all the classes that are mutually exclusive and have the same parent class should be included in the same BCT. As it was previously mentioned, multiple BCTs can derive from the same parent class (in such cases, only one of them will belong to the same tree as the parent class, and there will be a subsidiary tree for each of the rest of the BCTs).

In each BCT, the set of classes should be collectively exhaustive. In other words, all the instances from a given class *C* should belong to one of the classes from any BCT that is conditioned by such class *C*. If the classes from a given BCT are not collectively exhaustive, an exclusion class should be created.

These new classes could be negative classes with the form *no_[class_X]* for binary classification (e.g. *no_spam*) or residual classes with the form *other_[parent_class_Y]* for multiclass classification (e.g. *other_species*). At this point of the process, it is also possible to add new positive classes if needed, considering the data instances that would belong to the new classes, and relevant sources such as related literature or expert opinion.

It is important to mention that, in order to create a consistent taxonomy structure, it may be necessary to create auxiliary superclasses, like in the example from Fig. 6. These auxiliary superclasses can be dummy classes (e.g. the *(B/C)* class from the figure), or they can refer to actual concepts that are relevant to the classification problem but were not included in the initial set of classes for some reason.

Finally, if a BCT contains too many classes, it is possible to split it into multiple BCTs (see Fig. 9). To do so, group the classes based on their similarity and create one auxiliary superclass per group. After the split, instead of having one BCT with too many classes, you will have one BCT that contains the auxiliary superclasses, and each of these superclasses will condition a separate BCT containing the group of classes that corresponds to such superclass.

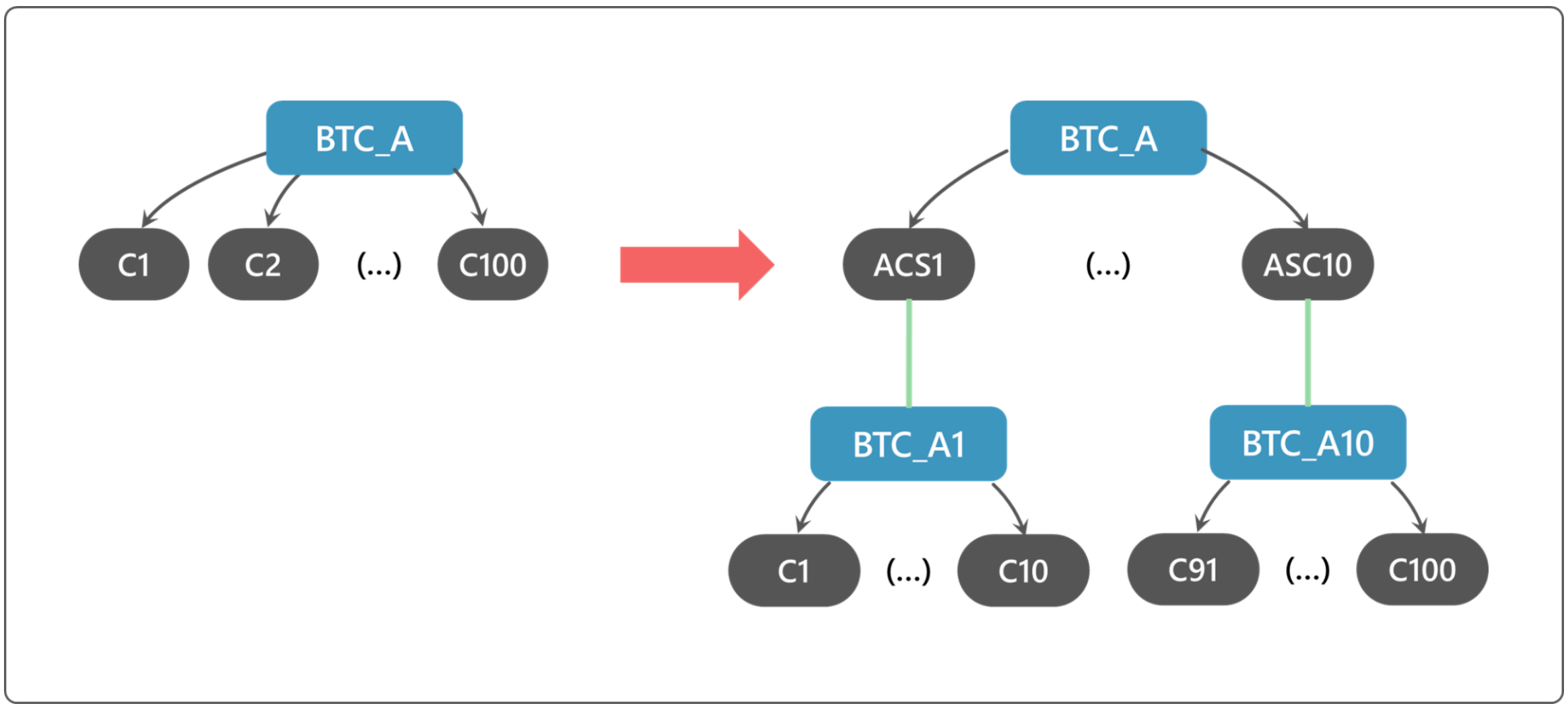

Fig. 9. A BCT with 100 classes is split into multiple BCTs, so that none of them contains more than 10 classes. *C1* to *C100* represent classes, and *ASC1* to *ASC10* represent auxiliary superclasses. "(...)" represent the parts of the tree that are not displayed for the sake of space.

4.1.4 Creation of the OWL file

Once the decision rainforest structure is defined, the next step is to generate an OWL 2 file based on it using Protégé (Musen, 2015). This process is straightforward, as the template provided in the project's repository (input_taxonomy_template.owx) can be used as a starting point, and the only requirement is to know how to add subclasses and relations. It is important to note that the created OWL file does not constitute the taxonomy on its own, as it contains not only the relevant classes but also some functional classes included for utilitarian purposes. The actual taxonomy file, containing only the classes and their corresponding logical axioms, is automatically generated during data preparation (see Section 4.2).

The decision rainforest includes the following functional classes:

- *taxonomy*: this class is the parent of all the tree names.
- tree names (e.g. *imaging_type*): there should be one initial class per class tree (both for main and subsidiary trees). This initial class will be considered to be the name of the tree. The classes that are descendent to these tree name classes are the ones that will be present in the actual taxonomy.
- *preprocessing*: this class is the parent to any class from the initial taxonomy that needs to be preprocessed (i.e. renamed, merged or split).
- *postprocessing*: this class is the parent to any compound class that should be added during postprocessing when a given combination of classes is present.

Apart from classes, the file contains the following properties:

- *has_subsidiary_tree*: this property is used to create relationships between classes and the subsidiary trees that are related to them.
- *maps_to*: any class from the original taxonomy that needs to be preprocessed (and therefore is a child class from *preprocessing*) should be related to the corresponding class or classes from the new taxonomy (descendants from the *taxonomy* class) through a *maps_to* relation. There are three possible preprocessing actions:
    - class renaming: the initial class should be related to the class in the new taxonomy with the correct name.
    - class merger: all the initial classes that need to be merged should be related to the corresponding class in the new taxonomy.
    - class split: the initial class should be related to the corresponding classes in the new taxonomy.
- *is_composed_by*: this property is used to create relationships between compound classes (descendants from the *postprocessing* class) and the corresponding taxonomy classes.

In Fig. 10, a decision rainforest and its associated OWL file are shown, both derived from a flat taxonomy containing seven classes. In the process, some of the initial classes are preprocessed: *us* is renamed as *ultrasound*, *roentgenogram* and *x_ray* are merged due to identical meaning, and the compound class *doppler_ultrasound* is split into the component classes *doppler* and *ultrasound*. An *is_composed_by* relation is created between *doppler_ultrasound* and its component classes to indicate that such compound class should be reintroduced during postprocessing. Finally, a *has_subsidiary_tree* relation was created between the class *ultrasound* and the tree name *attribute_doppler*.

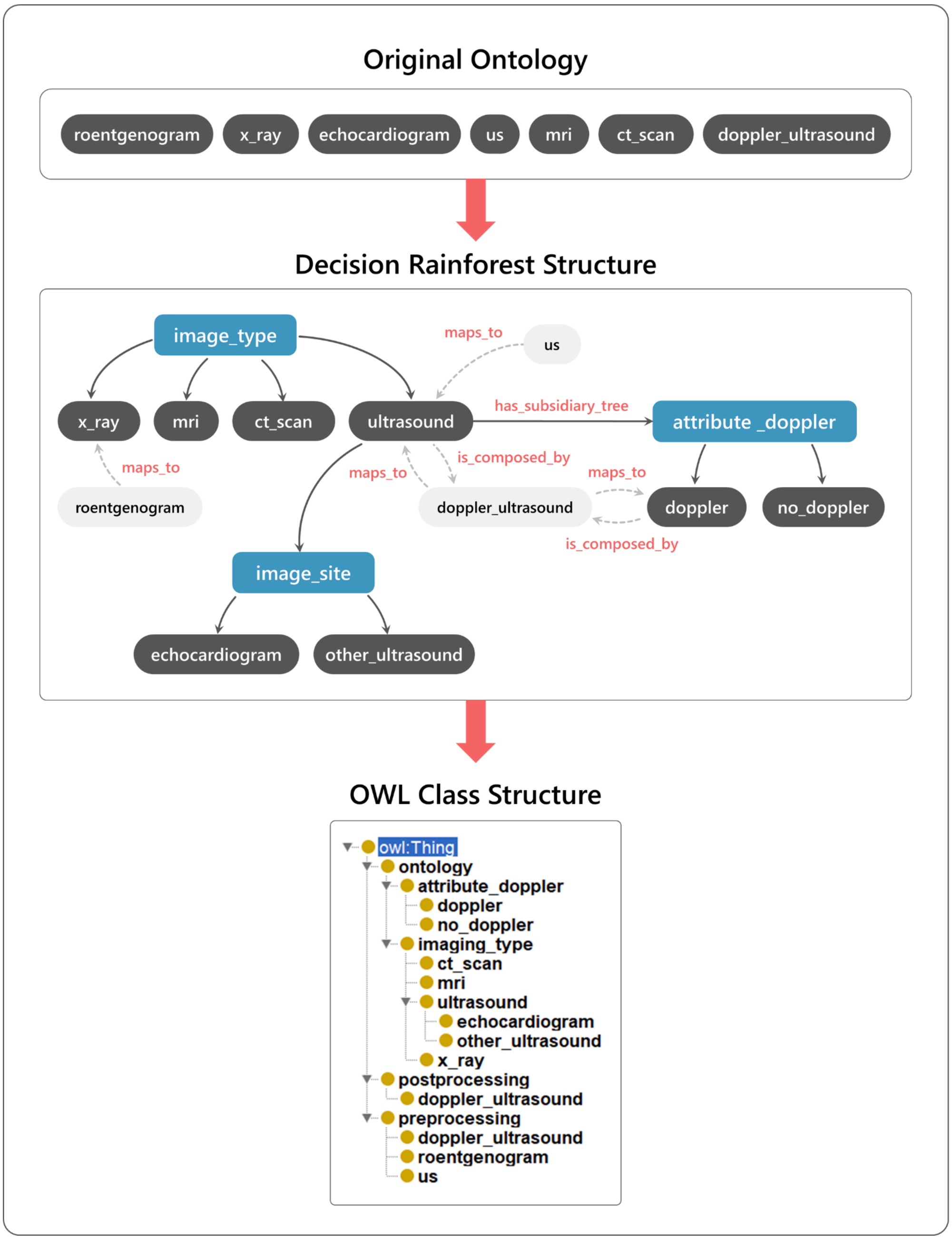

Fig. 10. A flat taxonomy is adapted into a decision rainforest structure, and an OWL file is created accordingly. In the decision rainforest structure, relations between actual classes (dark ovals) and the preprocessed or postprocessed classes (light ovals) are displayed. These relations are not displayed in the image of the OWL file due to space constraints.

### 4.1.5 Example of adaptation of a domain ontology

The purpose of this section is to demonstrate how domain ontologies can be adapted using the Multiplex Classification Framework. This task can be quite challenging, as ontologies that were not originally designed for ML usually contain several classes (including many compound classes) and their structure typically requires significant modifications. Additionally, class names or relationships may not always be self-explanatory, making it necessary to review documentation for accurate taxonomy adaptation. Moreover, it is crucial to recognize that larger and more complex ontologies are likely to include more domain-specific errors, as illustrated by cases such as SNOMED CT (Mortensen et al., 2014).

Figure 11 presents an example using a few classes from the clinical ontology SNOMED CT (SNOMED International, 2024a). For simplicity, some technical details were omitted (for instance, in reality not all the infections located in the lung are pneumonias). The adaptation was carried out following the steps outlined earlier:

- Class decomposition:
    - *viral_pneumonia* was split into *virus* + *lung*.
    - *infective_pneumonia* was split into *infection* + *lung*.
    - *respiratory_disease* was split into *respiratory_system* + *disease*.
- Identification of logical constraints:
    - *infection* is contained by *disease*, and *virus* is contained by *infection*.
    - *lung* is contained by *respiratory_system*.
    - *disease* also conditions *respiratory_system* and *lung* (because *disease* can be classified according to their finding site).
- Taxonomy structuring:
    - exclusion classes were created accordingly (e.g. *other_agent* was created to make classes from the BTC collectively exhaustive).
    - hierarchical relations were created between classes and one of their corresponding conditioned BTCs.
    - as *disease* has two conditioned BTCs, the subsidiary tree *finding_site* was created.

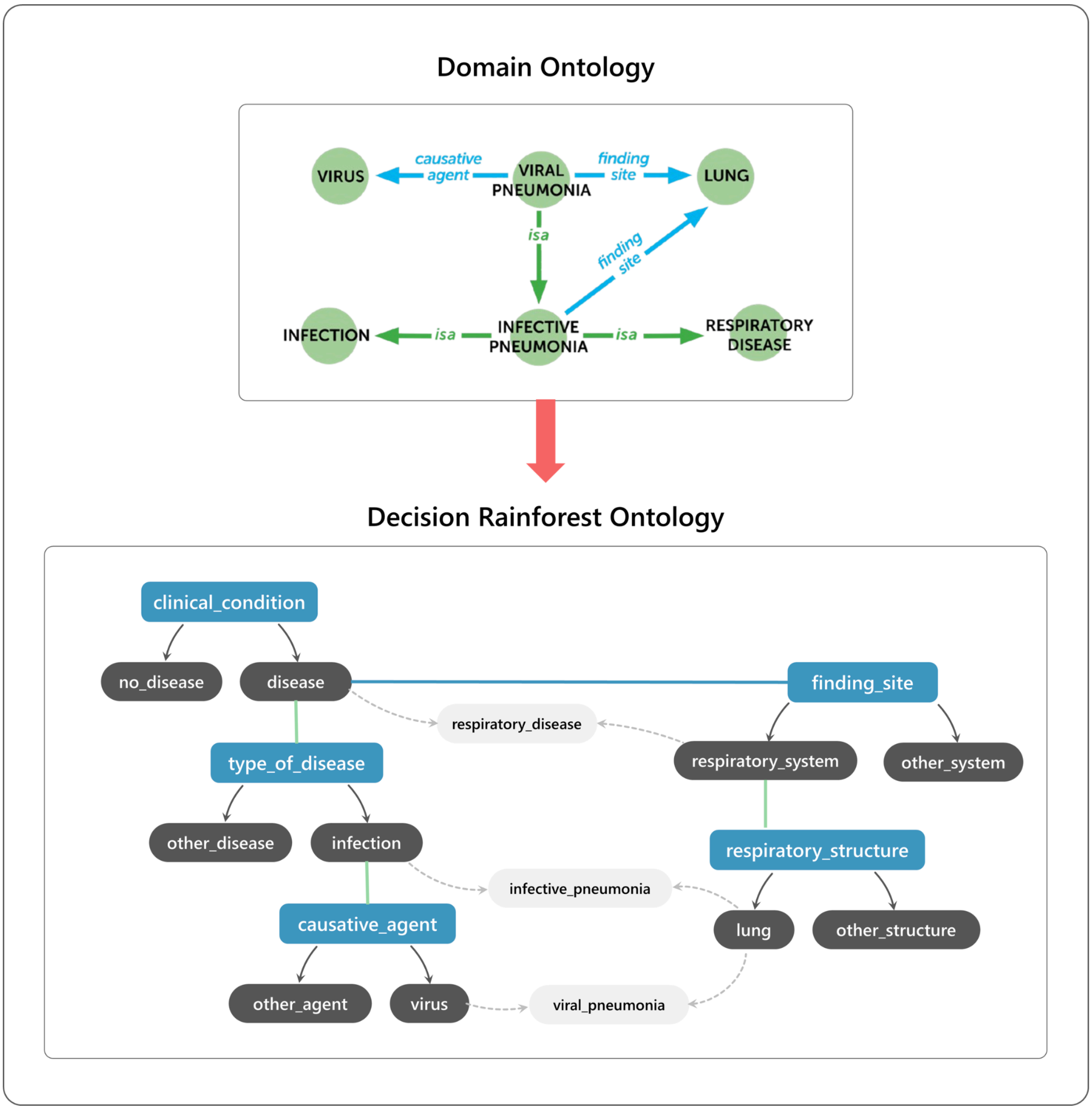

Fig. 11. Example of domain ontology adaptation. A subgraph of the SNOMED CT ontology (left) taken from an official ontology documentation (SNOMED International, 2024b), was transformed into a decision rainforest taxonomy (right). In the subgraph: green circles represent classes, and arrows represent relations; in the decision rainforest: blue rectangles represent BTCs, green lines represent hierarchical relations, blue lines represent subsidiary relations, black ovals represent classes, gray ovals represent compound classes, and gray arrows represent relations between compound classes and their corresponding component classes.

### *4.2 Data preparation*

Once the OWL file containing the decision rainforest is produced, adapting the dataset with annotations is straightforward: simply import the MultiplexDatasetProcessor class from the Multiplex Classification repository and instantiate it by providing the paths to the OWL file and the CSV file with the data as parameters. As an output, a CSV file will be generated with one column per model in the divergent cascading ensemble. In this section, a more detailed explanation of how the MultiplexDatasetProcessor works will be provided.

#### 4.2.1 Taxonomy processing

The first step in the data preparation process is taxonomy processing, which is carried out using the MultiplexTaxonomyProcessor class, leveraging the owlready2 library (Lamy, 2017). This class generates all necessary objects to adapt the dataset structure and produces a new OWL file with the output taxonomy. This new OWL file has the following characteristics:

- It only contains actual classes, whereas the input OWL file also includes functional classes (e.g., *preprocessing*).
- It has a tree structure with a single root class (called *'root_class'* by default), which is the ancestor of all the other classes.
- It has only one type of relation between classes: parent-child relations (both hierarchical and subsidiary relations from decision rainforests are transformed into this relation type).
- The corresponding axioms ensure that classes within the same BCT are disjoint unions of their parent class (i.e., they are mutually exclusive and collectively exhaustive).
- The corresponding class properties are created so that the new OWL file contains the same information as the input OWL file:
    - *tree_name*: name of the corresponding tree from the decision rainforest.
    - *class_path*: a string containing the class, all its ancestors, and their corresponding tree names.
    - *associated_compound_classes*: a list of compound classes associated with the given class (found under the *postprocessing* class from the decision rainforest).
    - *preprocessed_from*: a list of classes renamed, merged, or split during the creation of the decision rainforest and associated with the given class (found under the *preprocessing* class from the decision rainforest).

This new class structure is referred to as a Disjoint-Union-Based Tree (DUBT), as it represents a class tree where each class (except the root) belongs to a disjoint union set, and each parent class can be equivalent to multiple sets of disjoint union classes (see Fig. 12).

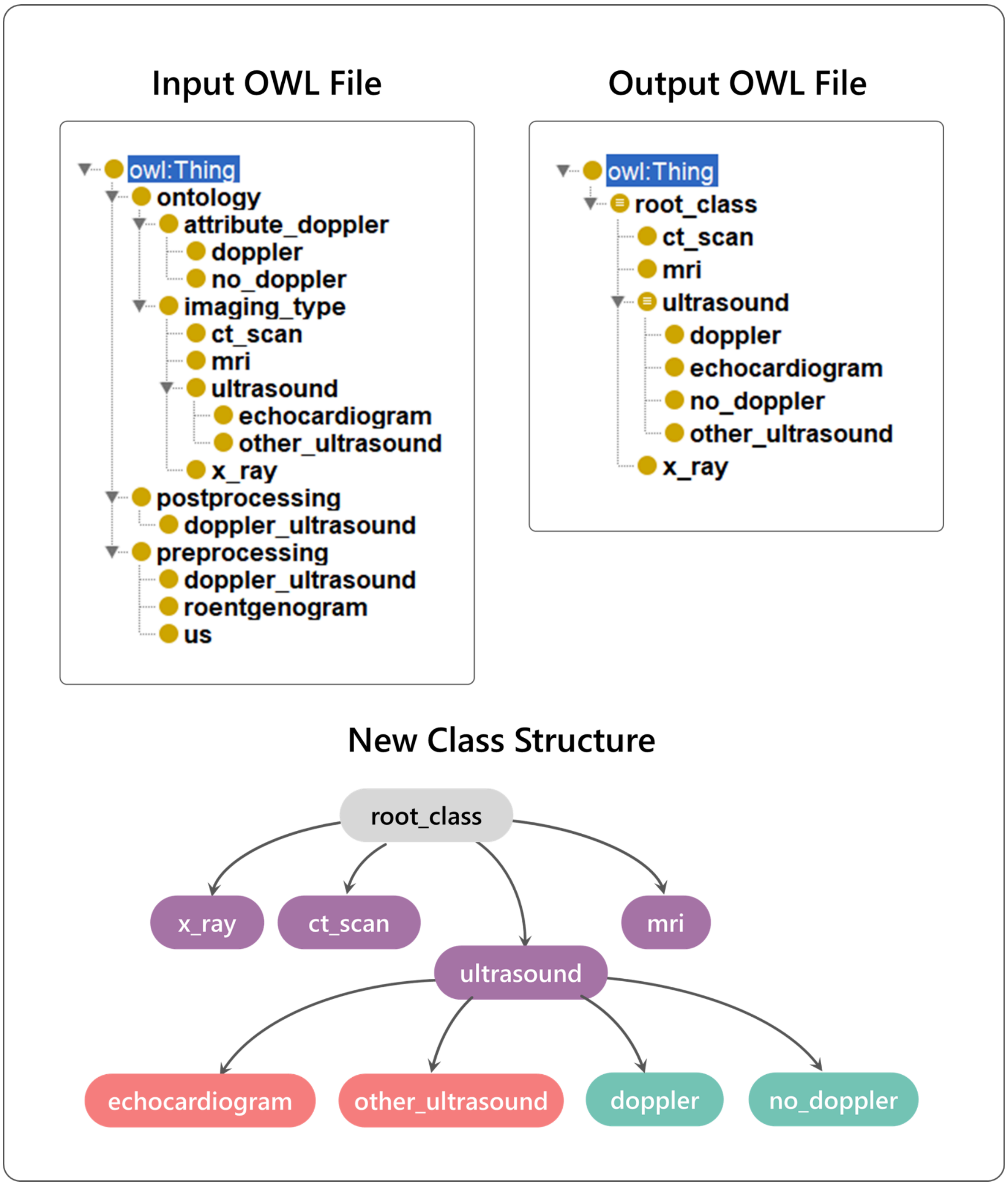

Fig. 12. Transformation of the decision rainforest taxonomy into a DUBT. The output OWL file includes axioms that define the corresponding disjoint union sets of classes (not shown in the image). Ovals in the class structure diagram represent classes, and classes of the same color belong to the same disjoint union set.

4.2.2 Quality of adapted taxonomies

The MultiplexTaxonomyProcessor class is designed to detect five types of errors that may be present in the input OWL file: repeated class names, empty class trees, single-child classes (no BCT should have only one class), graph structure (no class should have multiple parents), and recursive relations (no class should be their own ancestor).

The quality of a generic DUBT taxonomy was manually assessed based on its structure and axioms, as well as the pitfall catalog from the OOPS! methodology (Poveda-Villalón et al., 2014). A detailed explanation of this assessment can be found in Appendix A. Overall, the evaluation presented the following results:

- Most pitfalls are absent in a properly created DUBT taxonomy, mainly due to the error checks performed by the MultiplexTaxonomyProcessor and the absence of certain ontology elements prone to errors (e.g., no new relations or property chains are created).
- A few non-critical pitfalls are present:
    - One of these pitfalls is irrelevant to the scope of this framework (ML classification problems): "no OWL ontology declaration."
    - The other two are not actual pitfalls in our use case, as they are necessary for Multiplex Classification: "using miscellaneous classes" (a resource used to ensure BCT classes are collectively exhaustive) and "using ontology elements incorrectly" (in Multiplex taxonomies, elements typically treated as attributes in standard ontologies are considered classes because ML model predictions are always expressed as classes).
- For certain pitfalls, the outcome is "N/A" because the result depends on the specific user or classification problem (e.g., "missing annotations" is avoided if the user adds human-readable annotations to each class). However, none of these cases are relevant to the scope of this framework.

4.2.3 Data processing

The MultiplexDatasetProcessor is a class that leverages the Pandas library to transform an input CSV file with annotated data into the format needed for Multiplex classification considering the structure of the decision rainforest OWL file (The Pandas development team, 2024). For each instance, the labels in the input CSV file should be provided as a list in the same column (called 'label_list' by default). Any preprocessing actions defined in the decision rainforest OWL file (such as class renaming, merging, or splitting) are automatically applied by the MultiplexDatasetCreator. By default, the output is a CSV file with one column per model in the divergent cascading ensemble (see Fig. 13). The other two possible outcome formats are: 'multiplex_without_merging', which provides one column per BCT (without merging for multitask classifiers), and 'multilabel', where all labels are included in a single column as a list.

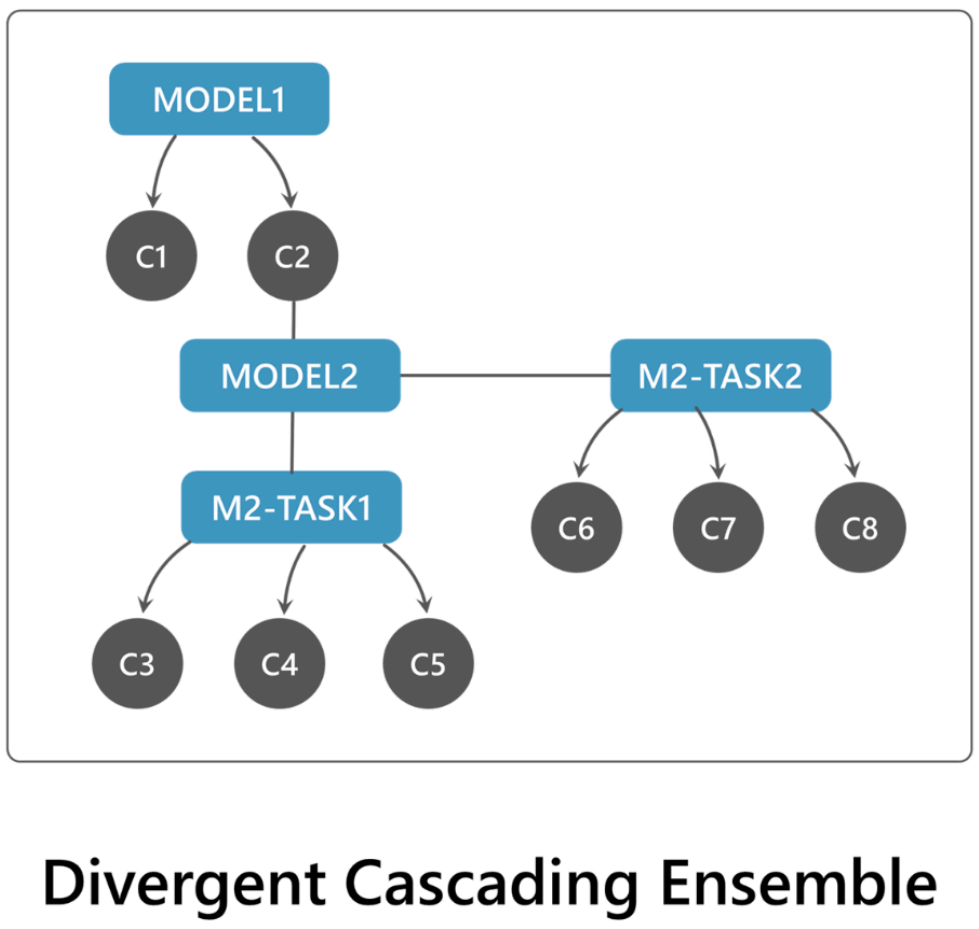

| id | MODEL1 | MODEL2 |
|---|---|---|
| id_000 | C1 | |
| id_001 | C2 | [C3,C7] |
| id_002 | C2 | [C5,C8] |
| id_003 | C1 | |
| id_004 | C2 | [C5,C6] |

Adapted Annotated Dataset

Fig. 13. Structure of the model ensemble (left) and the corresponding format of the adapted dataset (right). Labels from multitask models are included as lists within the same column.

#### 4.2.4 Improving data quality

During the data adaptation process, the quality of the dataset is improved by removing any set of incompatible labels while retaining labels that are not incompatible. For example, in a medical image classification problem, if a given data instance has the labels *radiology* (belonging to BCT_1) and *ultrasound* and *x_ray* (belonging to BCT_2) in the input CSV file, the output CSV file will only include the *radiology* label, since it does not conflict with the others.

In addition to removing label incompatibilities, the MultiplexDatasetProcessor class can enhance data quality by imputing missing labels using the “exclusion_classes” parameter. For example, if “exclusion_classes” = ['no_doppler'], any ultrasound instance without a label for the ‘doppler_attribute’ BCT will be assigned the label *no_doppler* by default.

### *4.3 Model ensembling*

#### 4.3.1 Model training

To develop a divergent cascading ensemble, it is necessary to train a set of ML submodels independently, where each submodel could be a binary classifier, a multiclass classifier, or a multitask classifier (refer to the Multiplex Classification repository for examples). This approach enables the fine-tuning of hyperparameters for each submodel independently, allowing better adaptation to the specific characteristics of each data subset. Additionally, modular approaches like this offer other documented benefits, such as partial reuse, simplified model maintenance, and enabling task divisions for collaborative efforts (Khan & Keet, 2015). It is worth noting that this modular model training does not require additional computational resources.

4.3.2 Model inference

At inference time, the outputs of the submodels are combined to generate the final predictions. This is done in a cascaded manner, where the inputs of the submodel corresponding to the BCT at the top of the tree are propagated downward —and, when necessary, divergently— based on the predicted classes. Using Fig. 2 as an example, if the initial model (a multiclass classifier) predicts that the input instance belongs to the class *x_ray*, then *x_ray* is the final prediction. However, if *ultrasound* is predicted, the input must also go through a second model (a multitask classifier) to obtain subsequent predictions (e.g., *echocardiogram* and *doppler*).

## 5. Experimental methods

Two experiments were conducted to compare model performance using a conventional approach (multiclass or multi-label) versus the Multiplex approach: the HyperKvasir experiment and the MultiCaRe experiment. The goal was to determine whether the Multiplex approach offers a significant improvement and to analyze the differences between the two approaches. In both experiments, the Multiplex taxonomies were created by adapting the original taxonomies from the datasets following the steps mentioned in Section 4.1. The same data preprocessing and training hyperparameters were used for both approaches to ensure that any difference in metrics could be attributed solely to the change in approach, rather than to hyperparameters or data quality issues.

Two different data sampling methods were tested: normal sampling, and optimized sampling (which consists of oversampling for conventional approaches, and a mix of normal sampling and oversampling for the Multiplex approach). In normal sampling, the model sees each image exactly once per epoch, following the standard distribution of data without any modification. In optimized sampling, the model can see each image anywhere from zero to multiple times, with the probability of sampling each image depending on the relative frequency of its class (images from underrepresented classes are more likely to be sampled).

### *5.1 The HyperKvasir experiment*

The HyperKvasir dataset is a manually annotated, multiclass image and video dataset for gastrointestinal endoscopy (Borgli et al., 2020). For this experiment, images with pathological finding labels were used, such as "*Esophagitis grade A*," "*Esophagitis grade B-D*," and "*Hemorrhoids*." The original taxonomy is flat (there are no hierarchies between classes) and includes 12 different classes. In this experiment, a BCT split was applied to transform the original classification problem into a sequence of simpler classification tasks.

### *5.2 The MultiCaRe experiment*

The MultiCaRe dataset is a multimodal case report dataset containing medical images that were programmatically labeled based on the content of their captions (Nievas Offidani & Delrieux, 2024). Images may have more than one label, making this a multi-label classification task. For this experiment, a subset of 31 classes was selected, including *endoscopy*, *colonoscopy*, and *mri*. The original taxonomy is flat, and it does not contain any logical constraints among the classes.

The Multiplex Classification Framework was applied to transform the original classification problem into a combination of simpler classification tasks. Before model training, compound classes were split, exclusion classes were added, and incompatible labels were filtered out (both for the multi-label model and the divergent cascading ensemble). In the training-test split, images from the same case report were kept within the same data subset.

The multi-label model's outputs were defined using confidence levels provided by the model and the same logical constraints from the Multiplex approach. Replacing threshold selection with this type of postprocessing is expected to improve the metrics of the conventional approach by avoiding label inconsistencies. This approach ensures that any differences in metrics between the two approaches are not due to logical inconsistencies in the outcome predictions. The postprocessing follows these steps: 1) selecting the class with the highest confidence score, 2) discarding mutually exclusive classes and selecting mutually inclusive ones, and 3) repeating the process until all classes are either discarded or selected. For example, consider the classes from Fig. 2, with the model predicting labels in the following order (from highest to lowest confidence): *ultrasound*, *x_ray*, *no_doppler*, *ct_scan*, *x_ray*, *doppler*, *echocardiogram*, *other_ultrasound*, *mri*. The process works as follows: first, the class with the highest confidence score (*ultrasound*) is assigned. Then, any incompatible classes (*x_ray*, *ct_scan*, and *mri*) are discarded, and any mutually inclusive classes are assigned (none in this case). The process is repeated for the remaining classes (*no_doppler*, *doppler*, *echocardiogram*, and *other_ultrasound*). The final outcome for this example would be: *ultrasound*, *no_doppler*, and *echocardiogram*.

## 6. Results

### *6.1 Adapted taxonomies*

In the HyperKvasir experiment, a BCT split was applied by grouping classes with related meanings, and creating auxiliary superclasses accordingly (see Fig. 14). This transformed the original 12-class multiclass classification problem into a multilevel classification problem with six BCTs: four binary, one with three classes, and one with six classes.

In the MultiCaRe experiment, the 31-class multi-label classification problem was reorganized into a set of seven BCTs: one with 13 classes, one with 7 classes, and the rest with no more than 4 classes (see Fig. 14). An auxiliary superclass, *other_staining*, was created to group low-prevalence pathology classes, addressing the issue of high class imbalance. Among the classes with *radiology* as the parent class, *angiography* was placed in a separate BCT because it is not mutually exclusive with the others (for instance, an image can be both *ct* and *angiography*). Since *radiology* conditions two BCTs, a multitask classifier was employed.

Both Multiplex taxonomies (in DUBT format) were tested in Protégé using OntoDebug with the HermiT reasoner (Schekotihin et al., 2019), and were confirmed to be coherent and consistent. These taxonomies have been uploaded to the BioPortal ontology repository: https://bioportal.bioontology.org/ontologies/MCO4HKE (containing the ontology from the HyperKvasir experiment) and https://bioportal.bioontology.org/ontologies/MCO4MCRE (which includes the ontology from the MultiCaRe experiment). It's important to mention that the ontologies presented in this work focus on the classification of medical images, an area that is still developing and holds significant potential for further exploration. Given the emerging nature of this domain, a lightweight ontology approach has been adopted, characterized by a smaller number of relations and instances compared to the more complex ontologies typically reviewed by this journal. This approach not only reflects the current state of knowledge in medical image classification but also facilitates future expansion and refinement as the field evolves. The simplicity of the ontology allows for better interoperability and applicability, without compromising its ability to provide meaningful value in the analysis and organization of medical images.

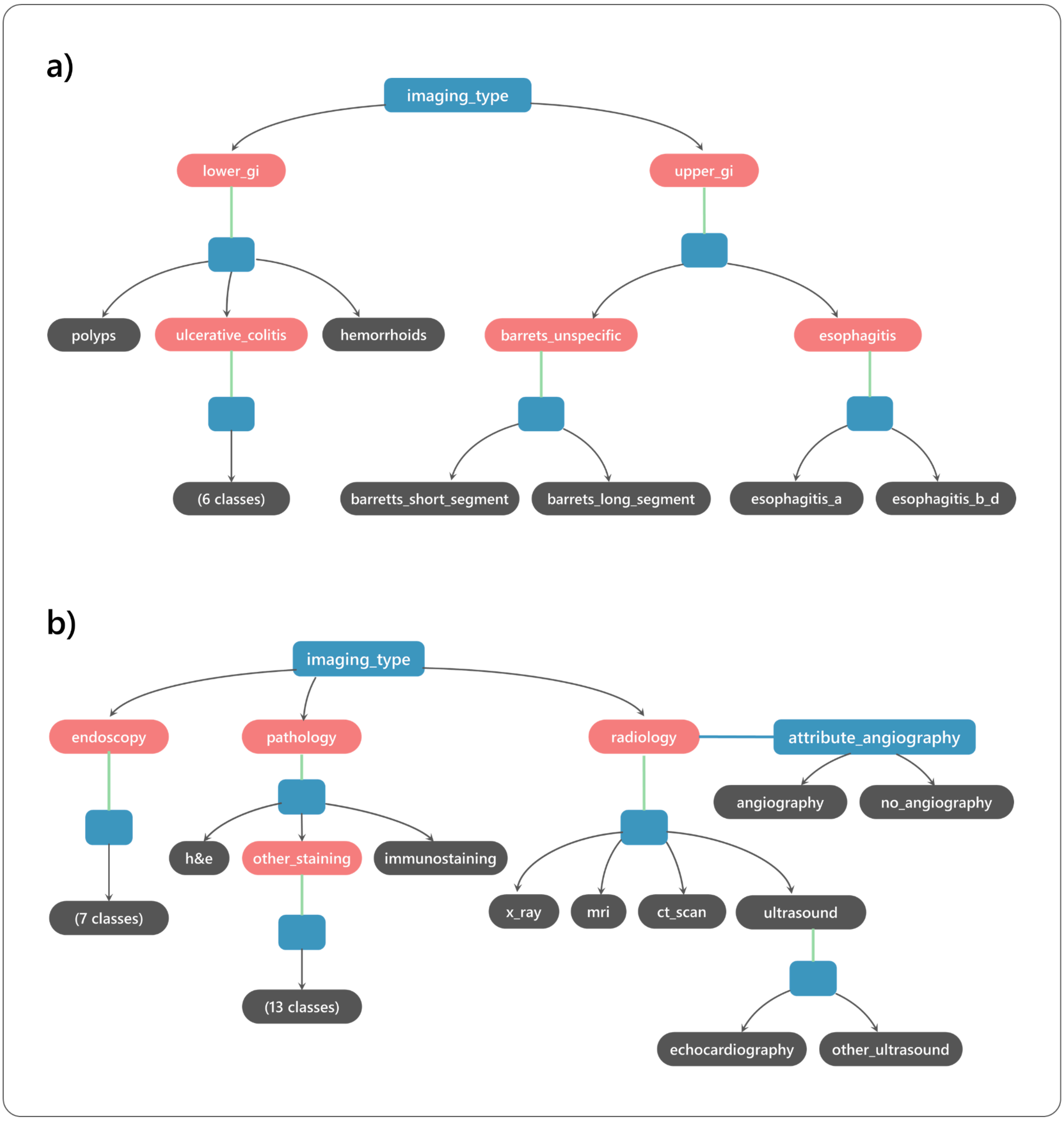

Fig. 14. Decision rainforests used in the Multiplex approach for the HyperKvasir experiment (a) and the MultiCaRe experiment (b). Blue rectangles represent BCTs, black ovals represent classes present in both the Multiplex and conventional approaches, and red ovals represent auxiliary superclasses. For the sake of space, only the name of the first BCT in each tree is displayed, and classes with multiple siblings are grouped together (for details on the taxonomy classes, refer to Appendix B).

### *6.2 Data quality improvement*

In the HyperKvasir experiment, no label inconsistency fixes were needed because each data instance had only one assigned label. In contrast, the MultiCaRe experiment revealed label inconsistencies in 5% of the instances (3,504 out of 64,799). These inconsistencies are explained by the fact that the dataset was not manually annotated (the labels were programmatically generated, as mentioned earlier). To improve data quality, these annotations were filtered out in both the multi-label and Multiplex approaches, with one key difference: the Multiplex approach allows to leverage data when incompatible labels share common ancestor classes (for example, if an image is labeled both as *ct* and *mri*, those labels are filtered out but the label corresponding to the auxiliary superclass *radiology* is kept).

### *6.3 Classification models*

Table 1 presents the results of both experiments, each using two sampling methods. In the HyperKvasir experiment, the Multiplex approach slightly outperformed the multiclass approach in both sampling methods, with a marginal difference of less than 4%. In contrast, the MultiCaRe experiment showed a more significant advantage for the Multiplex approach, with a difference of around 9%. A more detailed analysis (Fig. 15) reveals that lower class counts are associated with much higher F1-score gains, especially in normal sampling. The class-specific metrics for each experiment are provided in Appendix C.

Regarding training time, the MultiCaRe multi-label model took approximately 250 minutes to train, while the HyperKvasir multiclass model took 15 minutes. In both cases, the Multiplex counterparts required roughly twice as much time to train. In terms of storage space, each submodel required 308 MB: the MultiCaRe multi-label and HyperKvasir multiclass models each required 308 MB, while the MultiCaRe multiplex and HyperKvasir multiplex models required 1.848 GB each (308 MB x 6 submodels in both cases).

**Table 1.** Model Metrics.

| Experiment | Sampling | CA F1 | MA F1 | Abs. Diff. | % Diff. |
|---|---|---|---|---|---|
| HyperKvasir | Normal | 0.532 | 0.554 | 0.022 | 3.971 |
| HyperKvasir | Best | 0.554 | 0.559 | 0.005 | 0.894 |
| MultiCaRe | Normal | 0.476 | 0.533 | 0.057 | 10.694 |
| MultiCaRe | Best | 0.504 | 0.541 | 0.037 | 6.839 |

CA F1: F1-score from the conventional approach; MA F1: F1-score from the Multiplex approach; Abs. Diff.: MA F1 - CA F1; % Diff.: (Abs. Diff. / CA F1) * 100.

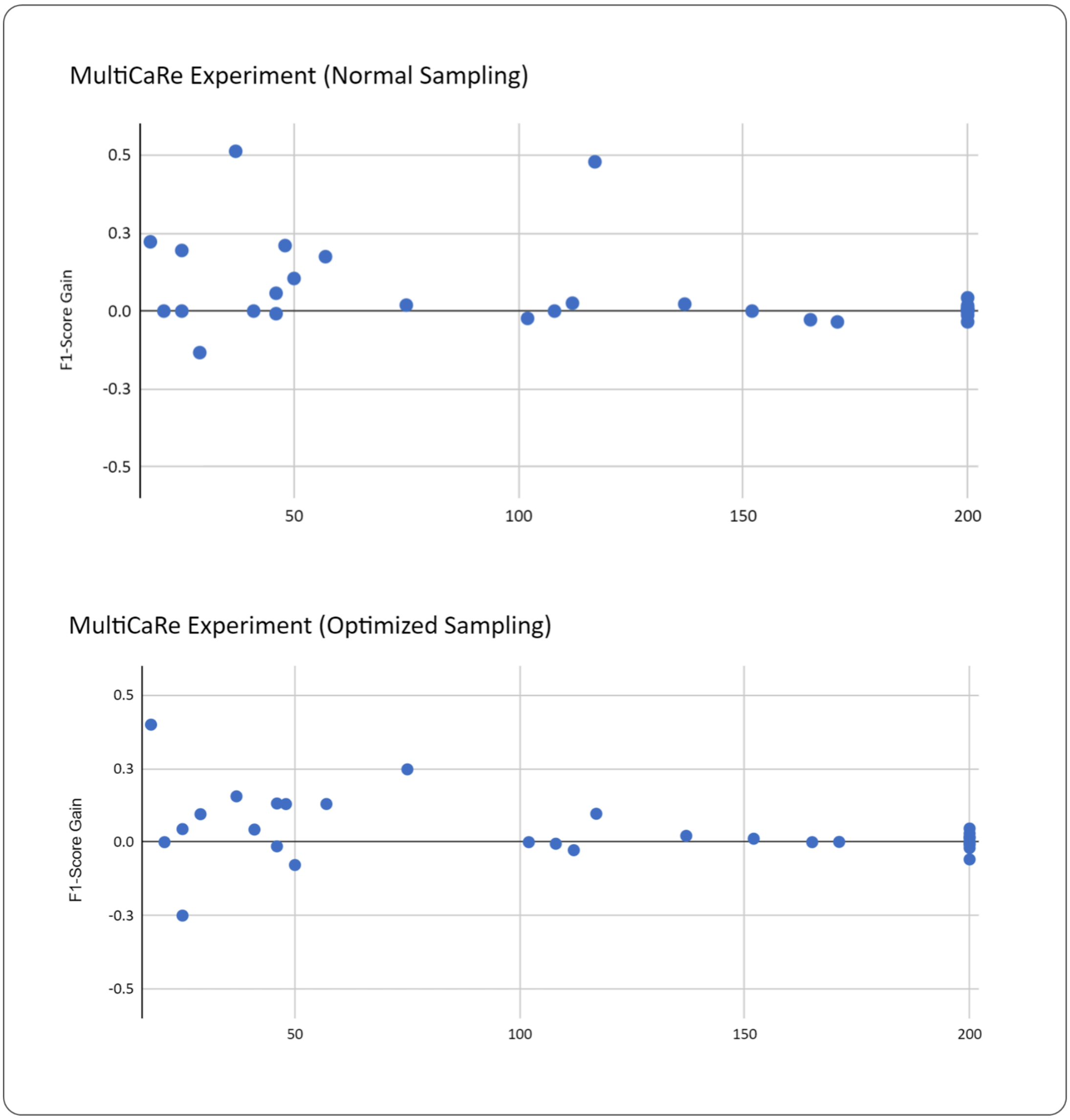

Fig. 15. Impact of class count on F1-score gain using normal sampling (upper) and optimized sampling (lower). Positive values indicate a higher F1-score with the Multiplex approach, while negative values indicate a higher F1-score with the multi-label approach. For display purposes, class counts above 200 were capped at that value.

## 7. Discussion

The results from our experiments demonstrate that using the Multiplex Classification Framework to adapt a classification problem can significantly impact the overall performance of classifiers in certain cases. The magnitude of this impact may be related to the total number of classes, the complexity of the problem, and the presence of classes with low counts. As demonstrated in Section 3.5, any conventional

ML classification problem can be transformed into Multiplex classification. Compared to other approaches, the Multiplex approach has the advantages and disadvantages listed below.

Main advantages:

- It supports any number of classes and logical constraints among them.
- It provides a novel method to address the class imbalance problem through BCT splitting.
- It eliminates the need for confidence threshold selection (common in multi-label classifiers) because each BCT always has only one predicted class.
- It adopts a modular structure, allowing, among other things, the independent fine-tuning of classifiers from different BCTs.
- It enables data quality improvement without requiring data review by removing incompatible labels, which may be particularly useful for silver-standard datasets.

Main disadvantages:

- It requires a deep understanding of the classification problem domain and experience in ontology engineering.
- It may not always provide a significant improvement in model metrics.
- It involves training a larger number of models, which can make the whole process more intricate

## 8. Conclusion

The Multiplex Classification Framework, introduced in this paper, presents a novel approach to addressing complex classification problems involving both sequential and simultaneous tasks. By leveraging problem transformation, ontology engineering, and model ensembling, the framework offers a versatile solution adaptable to a wide range of classification challenges. Our experimental results demonstrate that the Multiplex approach can lead to significant performance improvements, especially in cases with a large number of classes and class imbalances. Other advantages of this approach include its ability to accommodate any number of classes and logical constraints, its innovative method for addressing class imbalance, the elimination of confidence threshold selection, and its modular structure. However, the framework also has its limitations, as it may not always guarantee substantial metric improvements, it requires a deep understanding of the problem domain and expertise in ontology engineering, and it involves training multiple models, which can make the whole process more intricate.

In conclusion, the Multiplex Classification Framework offers a robust and flexible tool for addressing complex classification problems, particularly in scenarios involving numerous classes and logical constraints, such as medical image classification. Its modular design and innovative problem transformation techniques make it a valuable addition to the toolbox of researchers and practitioners in machine learning.

**Acknowledgments**

This research did not receive any specific grant from funding agencies in the public, commercial, or not-for-profit sectors.

**Appendix A: Ontology pitfall evaluation**

**Table 2.** Assessment of DUBT taxonomies based on the OOPS! methodology.

| Pitfall | Outcome | Comment |
|---|---|---|
| P01. Creating polysemous elements | OK | Not possible with the MultiplexTaxonomyProcessor. |
| P02. Creating synonyms as classes | OK | Completely overlapping classes are merged. |
| P03. Creating the relationship “is” instead of using ‘’rdfs:subClassOf”, ‘’rdf:type” or ‘’owl:sameAs” | N/A | No relation type is created. |
| P04. Creating unconnected ontology elements | OK | All created elements are connected. |
| P05. Defining wrong inverse relationships | N/A | No relation type is created. |
| P06. Including cycles in the hierarchy | OK | Not possible with the MultiplexTaxonomyProcessor. |
| P07. Merging different concepts in the same class | OK | Compound classes are split. |
| P08. Missing annotations | N/A | Adding human readable annotations to classes is outside the scope of the MultiplexTaxonomyProcessor. This should be done manually by the user. |
| P09. Missing basic information | OK | All the basic information for the classification task is present. |
| P10. Missing disjointness | OK | The correspondent disjoint axioms are present. |
| P11. Missing domain or range in properties | N/A | Object properties are not created |
| P12. Missing equivalent properties | N/A | This pitfall applies only to ontologies imported into another one. |
| P13. Missing inverse relationships | N/A | No relation type is created. |
| P14. Misusing ‘’owl:allValuesFrom” | N/A | “owl:allValuesFrom” is not used. |

**Table 2** (continuation).

| Pitfall | Outcome | Comment |
|---|---|---|
| P15. Misusing "not some" and "some not" | N/A | "not some" and "some not" are not used. |
| P16. Misusing primitive and defined classes | N/A | The only constraints on the classes are that they must be part of disjoint unions and each class must have a single parent. |
| P17. Specializing a hierarchy exceedingly | N/A | This aspect of the taxonomy depends on how the user defines the classification problem. |
| P18. Specifying the domain or range exceedingly | N/A | Domain and range are not defined. |
| P19. Swapping intersection and union | N/A | Domain and range are not defined. |
| P20. Misusing ontology annotations | OK | The contents of annotations are not misused. |
| P21. Using a miscellaneous class | NOT OK | Miscellaneous classes are present in order to ensure that the classes from a given classification task are collectively exhaustive (see Section 4.2.2). |
| P22. Using different naming criteria in the ontology | N/A | This depends on how the user names the classes. |
| P23. Using incorrectly ontology elements | NOT OK | Attributes are turned into classes because of how ML classification models work (see Section 4.2.2). |
| P24. Using recursive definition | OK | Not possible with the MultiplexTaxonomyProcessor. |
| P25. Defining a relationship inverse to itself | N/A | No relation type is created. |
| P26. Defining inverse relationships for a symmetric one | N/A | No relation type is created. |
| P27. Defining wrong equivalent relationships | N/A | No relation type is created. |
| P28. Defining wrong symmetric relationships | N/A | No relation type is created. |

**Table 2** (continuation).

| Pitfall | Outcome | Comment |
|---|---|---|
| P29. Defining wrong transitive relationships | N/A | No relation type is created. |
| P30. Missing equivalent classes | N/A | This pitfall applies only to ontologies imported into another one. |
| P31. Defining wrong equivalent classes | N/A | There are no equivalent classes. |
| P32. Several classes with the same label | OK | Not possible with the MultiplexTaxonomyProcessor. |
| P33. Creating a property chain with just one property | N/A | Property chains are not created. |
| P34. Untyped class | N/A | No resource is used as a class. |
| P35. Untyped property | N/A | No resource is used as a property. |
| P36. URI contains file extension | N/A | The definition of the URI and namespace is outside the scope of the MultiplexTaxonomyProcessor. |
| P37. Ontology not available | N/A | Taxonomies are not automatically available online when using the MultiplexTaxonomyProcessor. |
| P38. No OWL ontology declaration | NOT OK | No owl:Ontology tag is declared. |
| P39. Ambiguous namespace | N/A | The definition of the URI and namespace is outside the scope of the MultiplexTaxonomyProcessor. |
| P40. Namespace hijacking | N/A | The definition of the URI and namespace is outside the scope of the MultiplexTaxonomyProcessor. |

## Appendix B: Class references

**Table 3.** Classes from the HyperKvasir experiment (Multiplex approach).

| Class Name | Parent Class | Full Term |
|---|---|---|
| lower_gi | root_class | Lower gastrointestinal tract |
| upper_gi | root_class | Upper gastrointestinal tract |
| polyps | lower_gi | Polyps |
| ulcerative_colitis | lower_gi | Ulcerative colitis |
| hemorrhoids | lower_gi | Hemorrhoids |
| barrets_unspecific | upper_gi | Barrett's esophagus |
| esophagitis | upper_gi | Esophagitis |
| ulcerative_colitis_grade_0_1 | ulcerative_colitis | Ulcerative colitis (grade 0-1) |
| ulcerative_colitis_grade_1 | ulcerative_colitis | Ulcerative colitis (grade 1) |
| ulcerative_colitis_grade_1_2 | ulcerative_colitis | Ulcerative colitis (grade 1-2) |
| ulcerative_colitis_grade_2 | ulcerative_colitis | Ulcerative colitis (grade 2) |
| ulcerative_colitis_grade_2_3 | ulcerative_colitis | Ulcerative colitis (grade 2-3) |
| ulcerative_colitis_grade_3 | ulcerative_colitis | Ulcerative colitis (grade 3) |
| barretts_short_segment | barrets_unspecific | Barrett's esophagus (short segment) |
| barrets_long_segment | barrets_unspecific | Barrett's esophagus (long segment) |
| esophagitis_a | esophagitis | Esophagitis (grade A) |
| esophagitis_b_d | esophagitis | Esophagitis (grade B) |

**Table 4.** Classes from the MultiCaRe experiment (Multiplex approach).

| Class Name | Parent Class | Full Term |
|---|---|---|
| endoscopy | root_class | Endoscopy |
| pathology | root_class | Pathology |
| radiology | root_class | Radiology |
| arthroscopy | endoscopy | Arthroscopy |
| bronchoscopy | endoscopy | Bronchoscopy |
| colonoscopy | endoscopy | Colonoscopy |
| cystoscopy | endoscopy | Cystoscopy |
| egd | endoscopy | Esophagogastroduodenoscopy |
| gastroscopy | endoscopy | Gastroscopy |
| laryngoscopy | endoscopy | Laryngoscopy |
| h&e | pathology | Hematoxylin and eosin stain |
| immunostaining | pathology | Immunostaining |
| other_staining | pathology | Other staining |
| acid_fast | other_staining | Acid-fast staining |
| alcian_blue | other_staining | Alcian blue staining |
| congo_red | other_staining | Congo red staining |
| fish | other_staining | Fluorescence in situ hybridization |
| giemsa | other_staining | Giemsa staining |

**Table 4** (continuation).

| **Class Name** | **Parent Class** | **Full Term** |
|---|---|---|
| gram | other_staining | Gram staining |
| immunofluorescence | other_staining | Immunofluorescence |
| masson_trichrome | other_staining | Masson's trichrome staining |
| methenamine_silver | other_staining | Methenamine silver staining |
| methylene_blue | other_staining | Methylene blue staining |
| papanicolaou | other_staining | Papanicolaou staining |
| pas | other_staining | Periodic acid-Schiff staining |
| van_gieson | other_staining | Van Gieson's staining |
| ct | radiology | Computed tomography scan |
| mri | radiology | Magnetic resonance imaging |
| ultrasound | radiology | Ultrasound |
| x_ray | radiology | X-ray |
| angiography | radiology | Angiography |
| no_angiography | radiology | No angiography |
| echocardiogram | ultrasound | Echocardiogram |
| other_ultrasound | ultrasound | Other ultrasound |

## Appendix C: Class-specific metrics

**Table 5.** Class-specific metrics from the HyperKvasir experiment (normal sampling).

| Class | Training Counts | Test Counts | Multiclass F1 | Multiplex F1 | F1 Gain |
|---|---|---|---|---|---|
| barretts_long_segment | 33 | 8 | 0.3077 | 0.3636 | 0.0559 |
| barretts_short_segment | 43 | 10 | 0.4000 | 0.4444 | 0.0444 |
| esophagitis_a | 323 | 80 | 0.7898 | 0.7952 | 0.0054 |
| esophagitis_b_d | 208 | 52 | 0.7308 | 0.7429 | 0.0121 |
| hemorrhoids | 4 | 2 | 1.000 | 1.0000 | 0.0000 |
| polyps | 822 | 206 | 0.9831 | 0.9903 | 0.0072 |
| ulcerative_colitis_grade_0_1 | 27 | 8 | 0.1667 | 0.2667 | 0.1000 |
| ulcerative_colitis_grade_1 | 161 | 40 | 0.5205 | 0.5194 | -0.0011 |
| ulcerative_colitis_grade_1_2 | 9 | 2 | 0.0000 | 0.0000 | 0.0000 |
| ulcerative_colitis_grade_2 | 355 | 88 | 0.7513 | 0.7917 | 0.0404 |
| ulcerative_colitis_grade_2_3 | 22 | 6 | 0.0000 | 0.0000 | 0.0000 |
| ulcerative_colitis_grade_3 | 105 | 28 | 0.7407 | 0.7308 | -0.0099 |

**Table 6.** Class-specific metrics from the HyperKvasir experiment (optimized sampling).

| Class | Training Counts | Test Counts | Multiclass F1 | Multiplex F1 | F1 Gain |
|---|---|---|---|---|---|
| barretts_long_segment | 33 | 8 | 0.3333 | 0.4286 | 0.0953 |
| barretts_short_segment | 43 | 10 | 0.4348 | 0.3810 | -0.0538 |
| esophagitis_a | 323 | 80 | 0.8025 | 0.8025 | 0.0000 |
| esophagitis_b_d | 208 | 52 | 0.7407 | 0.7767 | 0.0360 |
| hemorrhoids | 4 | 2 | 1.0000 | 1.0000 | 0.0000 |
| polyps | 822 | 206 | 0.9734 | 0.9927 | 0.0193 |
| ulcerative_colitis_grade_0_1 | 27 | 8 | 0.3077 | 0.2667 | -0.0410 |
| ulcerative_colitis_grade_1 | 161 | 40 | 0.5647 | 0.5263 | -0.0384 |
| ulcerative_colitis_grade_1_2 | 9 | 2 | 0.0000 | 0.0000 | 0.0000 |
| ulcerative_colitis_grade_2 | 355 | 88 | 0.7416 | 0.7876 | 0.0460 |
| ulcerative_colitis_grade_2_3 | 22 | 6 | 0.0000 | 0.0000 | 0.0000 |
| ulcerative_colitis_grade_3 | 105 | 28 | 0.7458 | 0.7451 | -0.0007 |

**Table 7.** Class-specific metrics from the MultiCaRe experiment (normal sampling).

| Class | Training Counts | Test Counts | Multi-label F1 | Multiplex F1 | F1 Gain |
|---|---|---|---|---|---|
| acid_fast | 46 | 12 | 0.6400 | 0.6316 | -0.0084 |
| alcian_blue | 25 | 6 | 0.2500 | 0.4444 | 0.1944 |
| arthroscopy | 57 | 15 | 0.5455 | 0.7200 | 0.1745 |
| broncoscopy | 220 | 56 | 0.7037 | 0.7018 | -0.0020 |
| colonoscopy | 223 | 56 | 0.6182 | 0.6607 | 0.0425 |
| congo_red | 50 | 12 | 0.1176 | 0.2222 | 0.1046 |
| ct + angiography | 404 | 101 | 0.2069 | 0.1940 | -0.0129 |
| ct + not_angiography | 14873 | 3718 | 0.9426 | 0.9464 | 0.0038 |
| cystoscopy | 48 | 12 | 0.1429 | 0.3529 | 0.2101 |
| echocardiogram + angiography | 21 | 2 | 0.0000 | 0.0000 | 0.0000 |
| echocardiogram + not_angiography | 1186 | 296 | 0.8848 | 0.9020 | 0.0172 |
| egd | 112 | 28 | 0.4308 | 0.4561 | 0.0254 |
| fish | 165 | 41 | 0.6765 | 0.6486 | -0.0278 |
| gastroscopy | 41 | 11 | 0.0000 | 0.0000 | 0.0000 |
| giemsa | 171 | 43 | 0.6875 | 0.6526 | -0.0349 |
| gram | 75 | 19 | 0.5366 | 0.5556 | 0.0190 |
| h&e | 4862 | 1216 | 0.8960 | 0.8987 | 0.0026 |
| immunofluorescence | 152 | 38 | 0.8451 | 0.8451 | 0.0000 |
| immunostaining | 2524 | 630 | 0.8673 | 0.8701 | 0.0028 |
| laryngoscopy | 37 | 9 | 0.1538 | 0.6667 | 0.5128 |
| masson_trichrome | 108 | 28 | 0.5660 | 0.5660 | 0.0000 |
| methenamine_silver | 117 | 30 | 0.2353 | 0.7143 | 0.4790 |
| methylene_blue | 25 | 6 | 0.0000 | 0.0000 | 0.0000 |
| mri + angiography | 137 | 35 | 0.2041 | 0.2264 | 0.0223 |
| mri + not_angiography | 10794 | 2698 | 0.9485 | 0.9501 | 0.0016 |
| other_ultrasound + angiography | 18 | 4 | 0.0000 | 0.2222 | 0.2222 |
| other_ultrasound + not_angiography | 1949 | 488 | 0.8413 | 0.8517 | 0.0104 |
| papanicolaou | 102 | 26 | 0.6552 | 0.6316 | -0.0236 |
| pas | 258 | 65 | 0.4272 | 0.3925 | -0.0347 |
| van_gieson | 46 | 12 | 0.1429 | 0.2000 | 0.0571 |
| x_ray + angiography | 29 | 7 | 0.1333 | 0.0000 | -0.1333 |
| x_ray + not_angiography | 5566 | 1392 | 0.9374 | 0.9414 | 0.0040 |

**Table 8.** Class-specific metrics from the MultiCaRe experiment (optimized sampling).

| Class | Training Counts | Test Counts | Multi-label F1 | Multiplex F1 | F1 Gain |
|---|---|---|---|---|---|
| acid_fast | 46 | 12 | 0.5000 | 0.6316 | 0.1316 |
| alcian_blue | 25 | 6 | 0.4000 | 0.4444 | 0.0444 |
| arthroscopy | 57 | 15 | 0.5625 | 0.6923 | 0.1298 |
| broncoscopy | 220 | 56 | 0.7069 | 0.6481 | -0.0587 |
| colonoscopy | 223 | 56 | 0.6080 | 0.6372 | 0.0292 |
| congo_red | 50 | 12 | 0.3000 | 0.2222 | -0.0778 |
| ct + angiography | 404 | 101 | 0.1712 | 0.2176 | 0.0464 |
| ct + not_angiography | 14873 | 3718 | 0.9297 | 0.9243 | -0.0054 |
| cystoscopy | 48 | 12 | 0.3704 | 0.5000 | 0.1296 |
| echocardiogram + angiography | 21 | 2 | 0.0000 | 0.0000 | 0.0000 |
| echocardiogram + not_angiography | 1186 | 296 | 0.8784 | 0.8918 | 0.0134 |
| egd | 112 | 28 | 0.4528 | 0.4255 | -0.0273 |
| fish | 165 | 41 | 0.6486 | 0.6486 | 0.0000 |
| gastroscopy | 41 | 11 | 0.1000 | 0.1429 | 0.0429 |
| giemsa | 171 | 43 | 0.6517 | 0.6526 | 0.0009 |
| gram | 75 | 19 | 0.3077 | 0.5556 | 0.2479 |
| h&e | 4862 | 1216 | 0.8820 | 0.8987 | 0.0167 |
| immunofluorescence | 152 | 38 | 0.8333 | 0.8451 | 0.0117 |
| immunostaining | 2524 | 630 | 0.8524 | 0.8701 | 0.0176 |
| laryngoscopy | 37 | 9 | 0.3704 | 0.5263 | 0.1559 |
| masson_trichrome | 108 | 28 | 0.5714 | 0.5660 | -0.0054 |
| methenamine_silver | 117 | 30 | 0.6176 | 0.7143 | 0.0966 |
| methylene_blue | 25 | 6 | 0.2500 | 0.0000 | -0.2500 |
| mri + angiography | 137 | 35 | 0.2400 | 0.2615 | 0.0215 |
| mri + not_angiography | 10794 | 2698 | 0.9348 | 0.9150 | -0.0198 |
| other_ultrasound + angiography | 18 | 4 | 0.0000 | 0.4000 | 0.4000 |
| other_ultrasound + not_angiography | 1949 | 488 | 0.8395 | 0.8301 | -0.0094 |
| papanicolaou | 102 | 26 | 0.6316 | 0.6316 | 0.0000 |
| pas | 258 | 65 | 0.3934 | 0.3925 | -0.0009 |
| van_gieson | 46 | 12 | 0.2143 | 0.2000 | -0.0143 |
| x_ray + angiography | 29 | 7 | 0.0000 | 0.0952 | 0.0952 |
| x_ray + not_angiography | 5566 | 1392 | 0.9213 | 0.9243 | 0.0030 |